\pdfoutput=1

\documentclass[11pt]{article}

\usepackage{setspace}
\usepackage[]{misc/ACL2023}

\usepackage{amsmath,amssymb}
\usepackage{subcaption}
\usepackage{graphicx}
\usepackage{float}
\usepackage{stfloats}
\usepackage{fdsymbol}
\usepackage{pifont}
\usepackage[table]{xcolor}
\usepackage{adjustbox}
\usepackage{diagbox}
\usepackage{booktabs}
\usepackage{array}
\usepackage{booktabs}
\usepackage{multirow}
\usepackage{mdframed}
\usepackage{enumitem}
\usepackage{tabularx}

\usepackage{soul}
\usepackage[most]{tcolorbox}
\usepackage{cleveref}
\crefname{section}{§}{§§}
\Crefname{section}{§}{§§}

\definecolor{TodoColor}{rgb}{1,0.7,0.6}
\definecolor{TodoColor2}{HTML}{AACCAA}

\makeatletter\def\Hy@Warning#1{}\makeatother
\let\svthefootnote\thefootnote
\newcommand\blankfootnote[1]{%
  \let\thefootnote\relax\footnotetext{#1}%
  \let\thefootnote\svthefootnote%
}

\definecolor{exampleblue}{RGB}{40, 60, 170}
\definecolor{myyellow}{RGB}{251,230,153}
\definecolor{mygreen}{RGB}{169, 209, 142}  
\definecolor{myblue}{RGB}{157, 195, 230}
\definecolor{lightgray}{gray}{0.8}

\usepackage{times}
\usepackage{latexsym}
\usepackage{adjustbox}
\tcbuselibrary{listings,breakable}

\usepackage[T1]{fontenc}

\usepackage[utf8]{inputenc}

\usepackage{microtype}

\usepackage{inconsolata}

\usepackage[textwidth=2.25cm]{todonotes}
\usepackage{marginnote}

\definecolor{TodoColor}{rgb}{1,0.7,0.6}

\newcommand{\questionexample}[2]{
    \hspace{#1}
    \begin{minipage}{1.04\linewidth}
    \begin{quotebox}
        \setlength{\parindent}{0cm}
        #2
    \end{quotebox}
    \end{minipage}
}

\newcommand{\shading}[1]{%
  \tcbox[tcbox raise base, 
  left=0mm,right=0mm,top=0mm,bottom=0mm,boxsep=0pt,arc=0mm,
  boxrule=0pt,opacityfill=0.25,enhanced jigsaw,colback=gray!85!white,
  before=\relax,after=\relax]{\framebox[1.05\width]{\small \textsc{#1}}}
}

\newmdenv[
  linecolor=black,
  linewidth=1.2pt,
  topline=false,
  bottomline=false,
  rightline=false,
  innertopmargin=0mm,
  innerbottommargin=-0.5mm,
  skipabove=1.1\topsep,
  skipbelow=0.5\topsep,
]{quotebox}

\title{How to Engage Your Readers?$^{\vardiamondsuit}$ \\ Generating Guiding Questions to Promote Active Reading}

\author{
Peng Cui$^{1}$, Vilém Zouhar$^{1}$, Xiaoyu Zhang$^2$, Mrinmaya Sachan$^{1}$ \\
ETH Zürich Department of Computer Science$^{1}$, ETH AI Center$^{2}$ \\ 
\texttt{
peng.cui@inf.ethz.ch
}\\
}

\begin{document}
\maketitle

\begin{abstract}
Using questions in written text is an effective strategy to enhance readability.
However, what makes an active reading question good, what the linguistic role of these questions is, and what is their impact on human reading remains understudied. 
We introduce \textsc{GuidingQ}, a dataset of 10K in-text questions from textbooks and scientific articles.
By analyzing the dataset, we present a comprehensive understanding of the use, distribution, and linguistic characteristics of these questions.
Then, we explore various approaches to generate such questions using language models.
Our results highlight the importance of capturing inter-question relationships and the challenge of question position identification in generating these questions.
Finally, we conduct a human study to understand the implication of such questions on reading comprehension. 
We find that the generated questions are of high quality and are almost as effective as human-written questions in terms of improving readers' memorization and comprehension.
 \newline
 \newline
    \vspace{0.2em}
    \raisebox{-1mm}{
    \includegraphics[width=1.05em,height=1.05em]{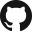}}
    \hspace{.5em}\parbox{\dimexpr\linewidth-2\fboxsep-2\fboxrule}
    {\small \href{https://github.com/eth-lre/engage-your-readers}{github.com/eth-lre/engage-your-readers}}
\end{abstract}

\section{Introduction}

Questions play an important role in reading comprehension. 
Through actively raising questions and seeking answers from the content during reading, readers can deeply engage with the text and achieve better comprehension \cite{bharuthram2017facilitating,syamsiah2018self}.
However, asking good questions is challenging and requires complex skills.

How can we facilitate readers' active thinking and questioning during reading?$^{\varheartsuit}$\footnote{Superscript symbols indicate the roles of questions in this paper, detailed in Section \ref{sec:role-schema}.}
An effective approach could be presenting valuable questions \emph{explicitly} in the text, which is a recognized strategy to enhance readability and engage readers \cite{haggan2004293}. 
An example is shown in Figure \ref{fig:intro}.
The writer first uses a title question (Q1) to arouse the interest of potential readers. 
Then, a group of questions (Q2-3) surfaces in the beginning to introduce the central topics to be explored.
In what follows, more questions (Q4-5) are raised and elaborated with the moving of discussion, holding the reader's attention throughout the reading.
From a linguistic perspective, these questions not only build up a coherent discourse structure \cite{curry2017questions}, but also serve as a communicative device that constructs a virtual dialogue between the writer and potential readers, thereby making the text more engaging and interactive \cite{hyland2002they}.

\begin{figure}[t]
\includegraphics[width=\linewidth]{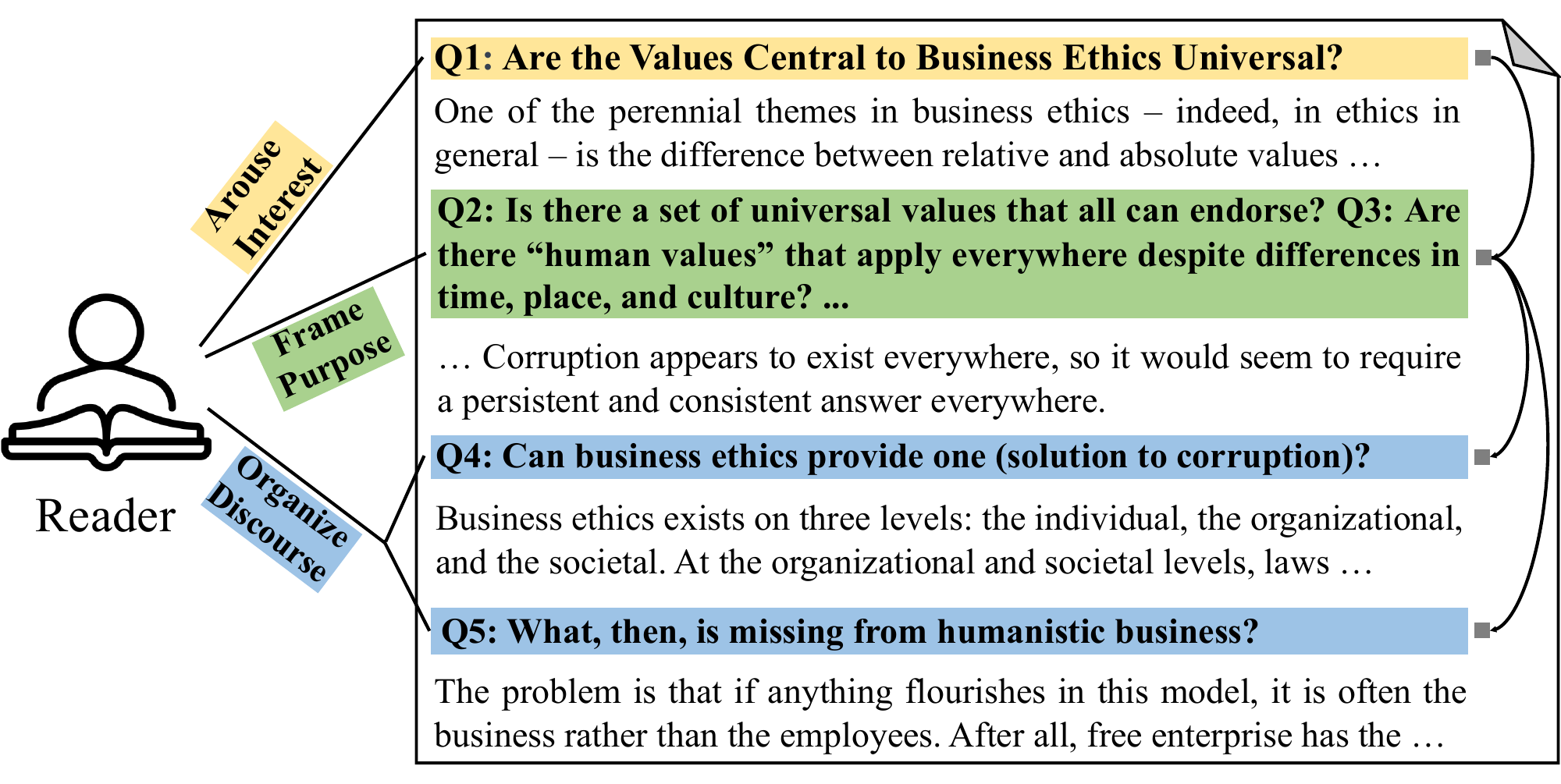}
\vspace*{-4mm}
\caption{We generate interconnected questions with \adjustbox{margin=1.5pt,bgcolor=myyellow}{diverse} \adjustbox{margin=1.5pt,bgcolor=mygreen}{rhetorical} \adjustbox{margin=1.5pt,bgcolor=myblue}{functions} \emph{during reading} to engage readers and improve comprehension.}
\vspace*{-4mm}
\end{figure}\label{fig:intro}

In this paper, we refer to these in-text questions as \emph{guiding questions}.
Despite being widely used, there is little understanding of the effect of such questions on human reading.
To fill this gap, 
we analyze how expert writers use guiding questions and explore how to model these questions with advanced language models.
Further, we hypothesize that these author-posed questions can complement and encourage readers' spontaneous self-questioning, thereby fulfilling the goal of active reading.
Based on these motivations, we
address 
the following research questions:
\begin{itemize}[left=0mm,noitemsep,topsep=1mm]
    \item \textbf{RQ1}: What is the use, distribution, and role of guiding questions in formal writing?$^{\clubsuit}$
    \item \textbf{RQ2}: How well can language models understand and generate guiding questions?$^{\clubsuit}$
    \item \textbf{RQ3}: What is the effect of these questions on human reading comprehension?$^{\clubsuit}$
\end{itemize}

To answer these questions, we start by curating \textsc{GuidingQ}, a dataset of 10,577 guiding questions from research articles and textbooks.
Since the two source texts are written by expert writers, we assume 
the questions are carefully designed to enhance readability and thus ideal for our research.
Through qualitative and quantitative analysis on \textsc{GuidingQ}, we summarize the question roles based on their \textbf{discourse} and \textbf{interactional} effects, and present their usage and distribution across the domains (RQ1, Sections \ref{sec:role-schema} and \ref{sec:data-construction}). 
Then, we explore various approaches to model these questions from three interrelated aspects (RQ2, Section \ref{sec:QG}): identifying question positions (where to ask), predicting question focus (what to ask), and finally generating the questions (how to ask).
Our results highlight the importance of inter-question relationships and the challenge of question position identification in generating guiding questions.

Finally, to validate whether and how the generated guiding questions can facilitate active reading (RQ3, Section \ref{Section-Human-Study}), we conduct a carefully designed human study where participants read articles with or without questions, complete a post-reading summarization test, and later evaluate the questions.
The results demonstrate that the generated questions are not only of high quality but can help readers produce better summaries, indicating an improved memory retention and understanding of the high-level information of the article.

\section{Related Work}
Here we focus on discussing the linguistic role of questions.
Questions can be used as a tool for discourse planning.
The Question Under Discussion (QUD) framework uses questions to interpret the discourse relationship between textual units within a document \cite{van1995discourse,roberts2012information,benz2017questions}.
For example, the relationship between \emph{"S$_a$: A night of largely peaceful protests ended early Monday in a bloody"} and \emph{"S$_b$: Hours earlier, Egypt's new interim leadership had narrowed
in on a compromise candidate to serve as the next prime minister."} can be described by question "\emph{What happened before the clash?}", where S$_a$, which elicits the question, is called anchor sentence and S$_b$ is the answer sentence.
Studying QUD with modern NLP techniques is a relatively new field where most efforts focus on data construction \cite{de-kuthy-etal-2018-qud,westera-etal-2020-ted, ko-etal-2022-discourse,wu-etal-2023-qudeval}.

While QUD provides a more flexible way to represent discourse connections compared to pre-defined fixed relationships (e.g., elaboration, condition) than other frameworks like Rhetorical Structured Theory (RST; \cite{mann1988rhetorical}), how to use these \emph{implicit} questions to assist readers is not straightforward.
In this paper, we instead focus on questions that are \emph{explicitly} presented in a text by the author, which we call guiding questions. 
We argue such questions are more powerful.
Besides their discourse effect, these questions directly interact with readers.
Therefore, they have the potential to influence readers' behaviors \cite{curry2017questions}. 
For example, articles titled with questions are found to obtain more downloads \cite{jamali2011article}. 

In terms of the goal, QUD strives to explore the dense space of questions to interpret exhaustive intra-document relationships, while we aim to distill sparse but crucial questions to engage readers without overwhelming them. 
We highlight the question position identification task as a preliminary step before generating questions. 
This dimension is neglected by previous QUD or generic Question Generation (QG) studies.
However, when questions are used as a communicative tool to interact with readers \emph{during} reading, when and where to raise them is arguably important. 
We hope this study sheds light on a greater understanding of the implication of questions in reading comprehension.
  
\begin{figure*}[ht]
    \centering
    \includegraphics[width=\textwidth]{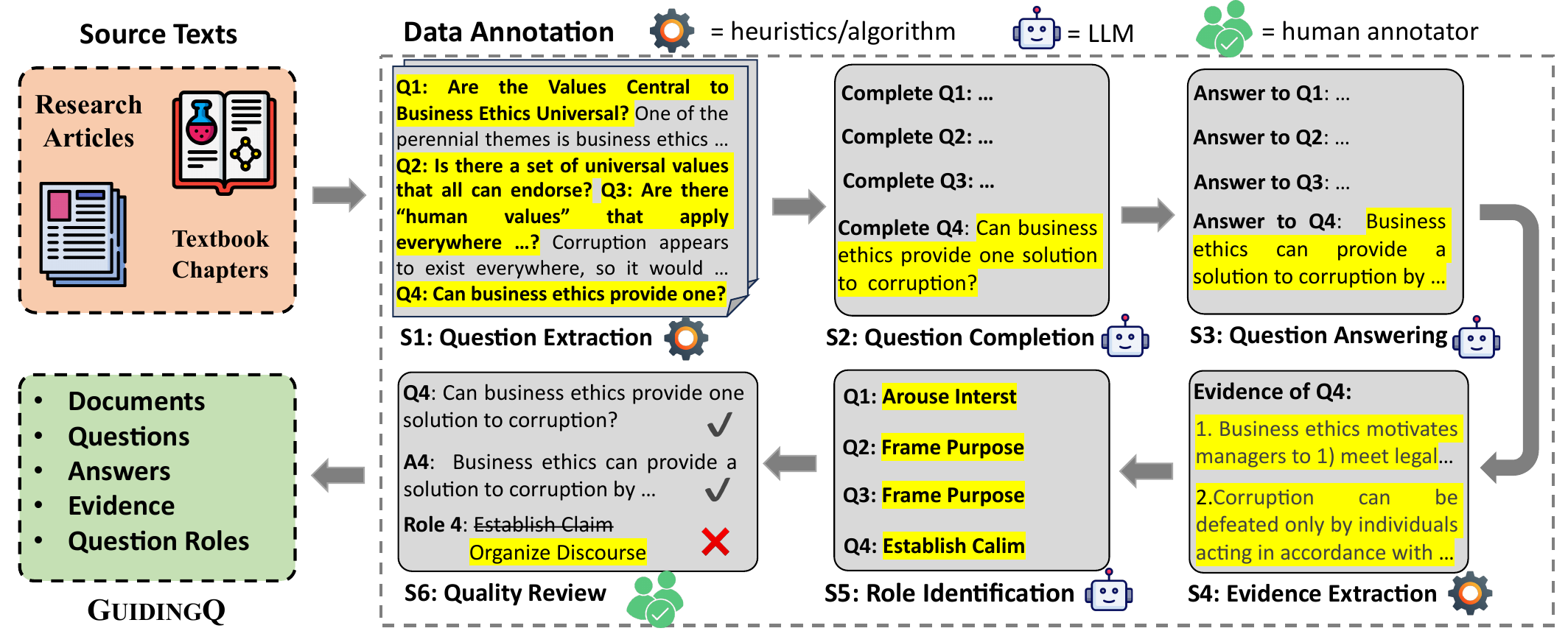}
    \vspace{-6mm}
    \caption{The construction pipeline of \textsc{GuidingQ}. Important outputs are \colorbox{yellow}{highlighted} }
    \label{fig:data-construction}
    \vspace{-4mm}
\end{figure*}
\section{Taxonomy of Guiding Questions} \label{sec:role-schema}
This section describes our taxonomy of guiding questions.
The taxonomy is built upon the discussion in \citet{hyland2002they}, which we adapt based on our dataset (Section \ref{sec:data-construction}).
In particular, we classify questions into five different roles based on their different discourse or interactional effects. 
Their definitions and examples are provided below.

\begin{itemize}[leftmargin=*,topsep=0mm]
    \item \textbf{Arouse Interest.$^{\vardiamondsuit}$}
    The first category refers to questions that appear in titles.
    Since the title is generally the reader's first encounter with a text, formulating it as an intuitive question can grab the reader's attention. 
    
    \questionexample{0mm}{
    \small
    \textbf{\texttt{Questions in titles:}}  \\
    \emph{\textcolor{exampleblue}{How do Philosophers arrive at truth?}} \\
    \emph{\textcolor{exampleblue}{Is there no quantum form of Einstein Gravity?}} \\
    \emph{\textcolor{exampleblue}{Why do house-hunting ants recruit in both directions?}} 
    }
    
    Note that although the title question usually reveals the main topic of an article, the breadth of the content is not always encapsulated by it
    \item \textbf{Frame Purpose.$^{\clubsuit}$}
    This type of question often surfaces and clusters in the beginning section to foreground the central topics to be explored. \vspace{2mm}

    \questionexample{0mm}{
    \small
    Several aspects of this theory need further investigation. 
    \emph{\textcolor{exampleblue}{Is it possible to achieve predictable refractive changes? Can this be achieved through an intact epithelium?}} ... This paper describes the use of a novel device ...
    }
    




    Writers pose these questions to provide an agenda for the article and then pick them up again in later sections to direct readers through the reading.
    \item \textbf{Organize Discourse.$^{\spadesuit}$}
    Questions can also serve as \emph{subheadings} to structure the text, guiding readers by explicitly introducing shifts in information and identifying what will be discussed in the ensuing section.
    
    \questionexample{0mm}{
    \small
    \emph{\textcolor{exampleblue}{What are the advances of telecommuting?}} The term telecommuting emerged in the 1970s to ... \emph{\textcolor{exampleblue}{What are the drawbacks of telecommuting?}} In 2013, Yahoo’s then-CEO, Marissa Mayer, ended ... \emph{\textcolor{exampleblue}{What are the ethical challenges of telecommuting?}} ...}
    
    Noticeably, such questions usually appear multiple times throughout an article, collectively creating a sense of progression toward a greater understanding of the topic.    
    \item \textbf{Establish Claim.$^{\varheartsuit}$} 
    Another use of questions is to introduce and emphasize the writer's arguments rather than to seek the reader's interaction or viewpoint. 
    
    \questionexample{0mm}{
    \small
    \emph{\textcolor{exampleblue}{What contributes to a corporation’s positive image over the long term?}} Many factors contribute, including \ul{a reputation for treating customers and employees fairly and for engaging in business honestly.}}

    A distinct feature of such questions is that the writer often provides a clear \ul{answer} (i.e., the argument), usually close to the question, thereby limiting the reader's alternative interpretations to the preferred one.
    \item \textbf{Provoke Thought.$^{\bigstar}$} 
    Finally, there are some ``genuine'' questions that do not anticipate specific responses within the text. 
    Therefore, they can facilitate the reader's active thinking to the greatest extent.
    
    \questionexample{0mm}{
    \small
    If the technological resources of today’s governments had been available to the East Germany Stasi and the Romanian Securitate, \emph{\textcolor{exampleblue}{would those repressive regimes have fallen? How much privacy and freedom should citizens sacrifice to feel safe?}} \textbf{\texttt{{[}END{]}}}
    }

\end{itemize}
It's worth noting that different roles are not mutually exclusive.
For instance, an \shading{Arouse Interest} question may also provoke thoughts and vice versa. 
Nevertheless, we focus on understanding the main role of a question. 

\section{\textsc{GuidingQ Dataset}} \label{sec:data-construction}

In this section, we first discuss the choice and rationale of source texts (\cref{data-source}), followed by the construction pipeline (Figure \ref{fig:data-construction}) of the \textsc{GuidingQ} dataset (\cref{data-annotation}).
Then, we present a series of distributional features of guiding questions (\cref{subsec:data_analysis}).

\subsection{Source Texts} \label{data-source}
We select \textbf{scientific articles} and \textbf{textbooks} as the source texts to build the dataset.
Our choice is based on two considerations.
First, their writer-reader discourses have a clear communicative intent,
either peer-to-peer or teacher-to-student, which can motivate the use of questions. 
Second, they are formal texts written by experts, ensuring that questions presented in them are strategically used to enhance readability.
Specifically, we use textbook chapters collected from an online free publisher OpenStax\footnote{\href{https://openstax.org/}{openstax.org}} \cite{singh-etal-2023-enhancing} and research papers from the arXiv and PubMed datasets \cite{cohan-etal-2018-discourse}. 

\subsection{Construction Pipeline} \label{data-annotation}
We describe the main steps of collecting and annotating \textsc{GuidingQ} below.
\vspace{1mm}

\noindent \textbf{S1: Question Extraction.} 
We start by extracting questions from source texts by detecting interrogative marks.
We only keep documents with at least three questions, indicating the writer actively used questions in the writing.
\vspace{1mm}

\noindent \textbf{S2: Question Completion.} 
Since the extracted questions are a part of the source texts, they are not always semantically complete due to omissions or unclear pronouns, e.g., \emph{What central point might constitute \underline{such a code}?}
Therefore, we first identify and complete such questions based on the context.
We do this because it is the first step to understanding the meaning of such questions.
\vspace{1mm}

\noindent \textbf{S3: Question Answering.}
Next, we generate the answer to each question.
In particular, the answer should be detailed enough and solely based on the article.
Therefore, we use the article's words as the answer whenever possible. 
If a question is not discussed in the article (e.g., \shading{Provoke Thoughts} question), we label it as "\texttt{NO ANSWER}."
\vspace{1mm}

\begin{table}[t]
\small
\centering
\begin{tabular}{@{}lcc@{}}
\toprule
\bf GuidingQ                 & \multicolumn{1}{l}{\bf Textbook} & \multicolumn{1}{l}{\bf Research Articles} \\ \midrule
\# documents             & 621                          & 1,501                                \\
\# avg. words /doc.      & 2,404                        & 2,140                               \\
\# questions             & 3,593                        & 6,964                               \\
\# avg. words / question & 13.8                         & 21.0                                \\
\# avg. questions / doc. & 5.79                         & 4.63                                \\ \bottomrule
\end{tabular}
\caption{Statistics of the \textsc{GuidingQ} dataset}
\label{tab:stats}
\vspace{-4mm}
\end{table}

\noindent \textbf{S4: Evidence Extraction.} 
Given a produced answer, we automatically extract \emph{supporting sentences} from the article as evidence.
We do this by greedily searching a set of sentences that has the maximum Rouge score with the answer, which is the standard way to find Oracle sentences for extractive summarization systems \cite{nallapati2017summarunner}.
For questions without answers as per \textbf{S3}, we directly label them as "\texttt{NO EVIDENCE}."
\vspace{1.5mm}

\noindent \textbf{S5: Question Role Identification.}
Finally, we identify the role of each question.
We make this the last step as the information collected in previous steps (complete question, answer, evidence) can help in understanding the role of a question. 

Recent studies have shown that LLMs with carefully designed prompts are comparable to or even better than human annotators \cite{he2023annollm,zhang-etal-2023-llmaaa,tornberg2023chatgpt}.
To reduce human effort and enable data scaling in the future, we use ChatGPT to complete the annotation steps \textbf{S2, S3,} and \textbf{S5}. 
See prompts in Tables \ref{tab:ann_prompt_qc}, \ref{tab:ann_prompt_qa}, and \ref{tab:ann_prompt_qri}.

\vspace{1.5mm}

\noindent \textbf{S6: Quality Control.} 
We adopt a Human-LLM co-annotation paradigm \cite{li-etal-2023-coannotating}.
Concretely, we ask the model to provide a confidence level for each of its outputs. 
Examples below a certain level are reviewed by human annotators, followed by an overall quality review, detailed in Appendix \ref{appendix:details-annotation}.

\begin{figure}[t]
    \centering
    \includegraphics[width=\columnwidth]{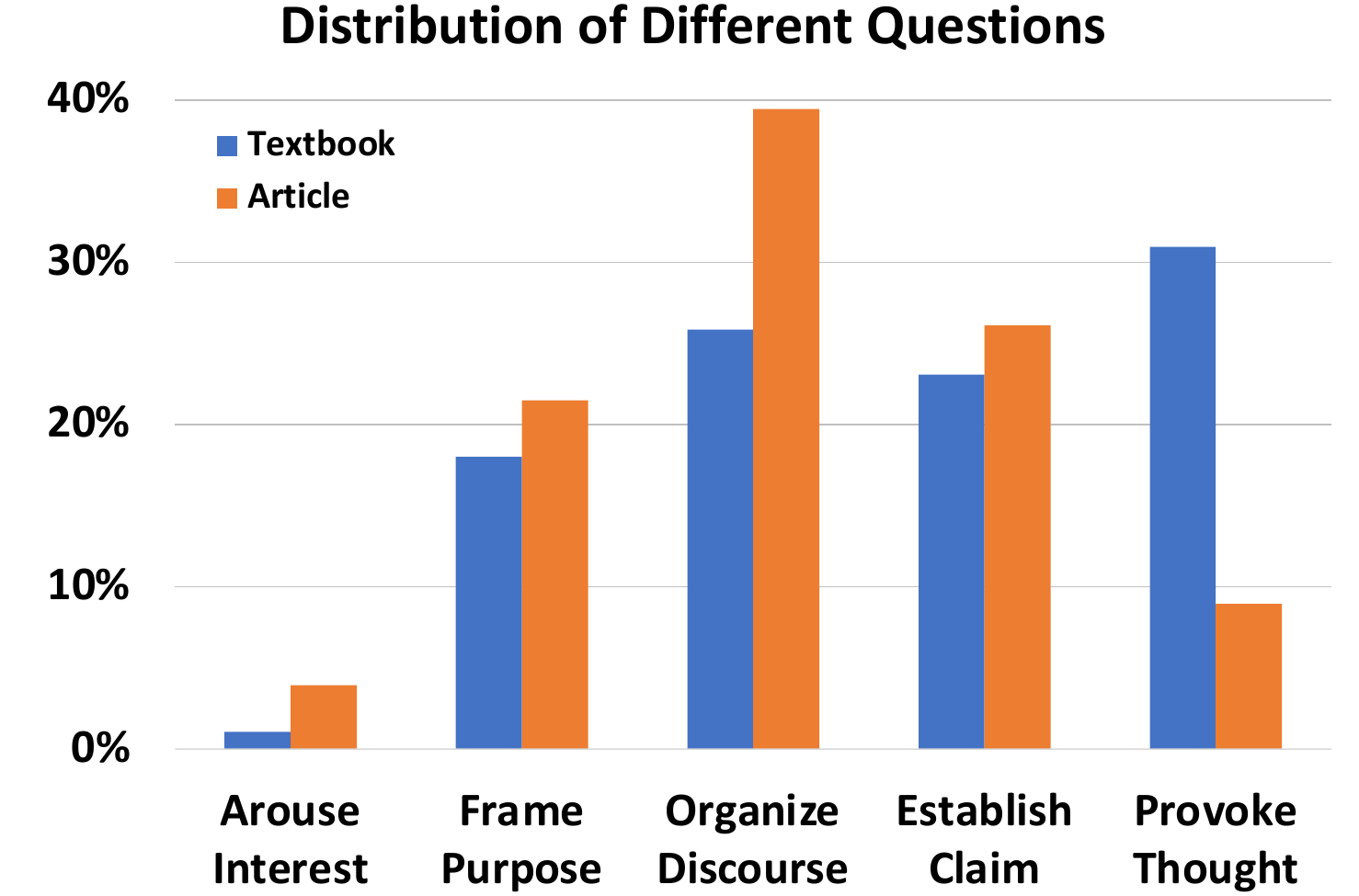}
    \caption{Distribution of different question roles.}
    \label{fig:type-distribution}
    \vspace*{-4mm}
\end{figure}
\subsection{\textsc{GuidingQ} Analysis} \label{subsec:data_analysis}
The statistics of the \textsc{GuidingQ} dataset is summarized in Table \ref{tab:stats}.
In general, textbook chapters use questions slightly more frequently than research articles.
This is possibly because teacher-to-student interactions are more inclined to involve questions than peer-to-peer ones.

\paragraph{What is the distribution of question roles?$^{\spadesuit}$}
In \Cref{fig:type-distribution}, we compare the distributions of different questions on the two subsets. 
As can be seen, research articles contain more \shading{Frame Purpose} and \shading{Organize Discourse} questions, while textbooks favor \shading{Provoke Thought} questions possibly due to their educational purpose.
Besides, \shading{Arouse Interest} (title) questions are more common in articles than textbooks, although both proportions are small.
A possible reason is that research articles exist in a competitive environment where potential readers are confronted with a large number of papers, under which circumstances interrogative titles could help attract readers \cite{haggan2004293,jamali2011article}. 
\vspace{1mm}

\noindent \textbf{How diversely do human writers use guiding questions?$^{\spadesuit}$} 
To understand this, we measure the number of question roles used in an article.
The results are shown in Table \ref{tab:num_unique_q}.
Since research articles contain fewer questions per article, their question roles are slightly less diverse than those in textbooks. 
Overall, the majority of articles (70\%+) use less than three types of questions, suggesting room for improvement in human questioning strategies.
\begin{table}[htbp]
\centering
\small
\begin{tabular}{@{}lccccc@{}}
\toprule
\bf \# Question Types   & \bf 1    & \bf 2    & \bf 3    & \bf 4   & \bf 5   \\ \midrule
Textbooks         & 21.1 & 44.8 & 26.8 & 7.0 & 0.2 \\
Articles & 35.2 & 45.5 & 17.2 & 2.0 & <0.1 \\ \bottomrule
\end{tabular}
\caption{Distribution (\%) of the number of unique question roles in one article.}
\label{tab:num_unique_q}
\vspace{-3mm}
\end{table}
\begin{figure}[t]
    \centering
    \includegraphics[width=\columnwidth]{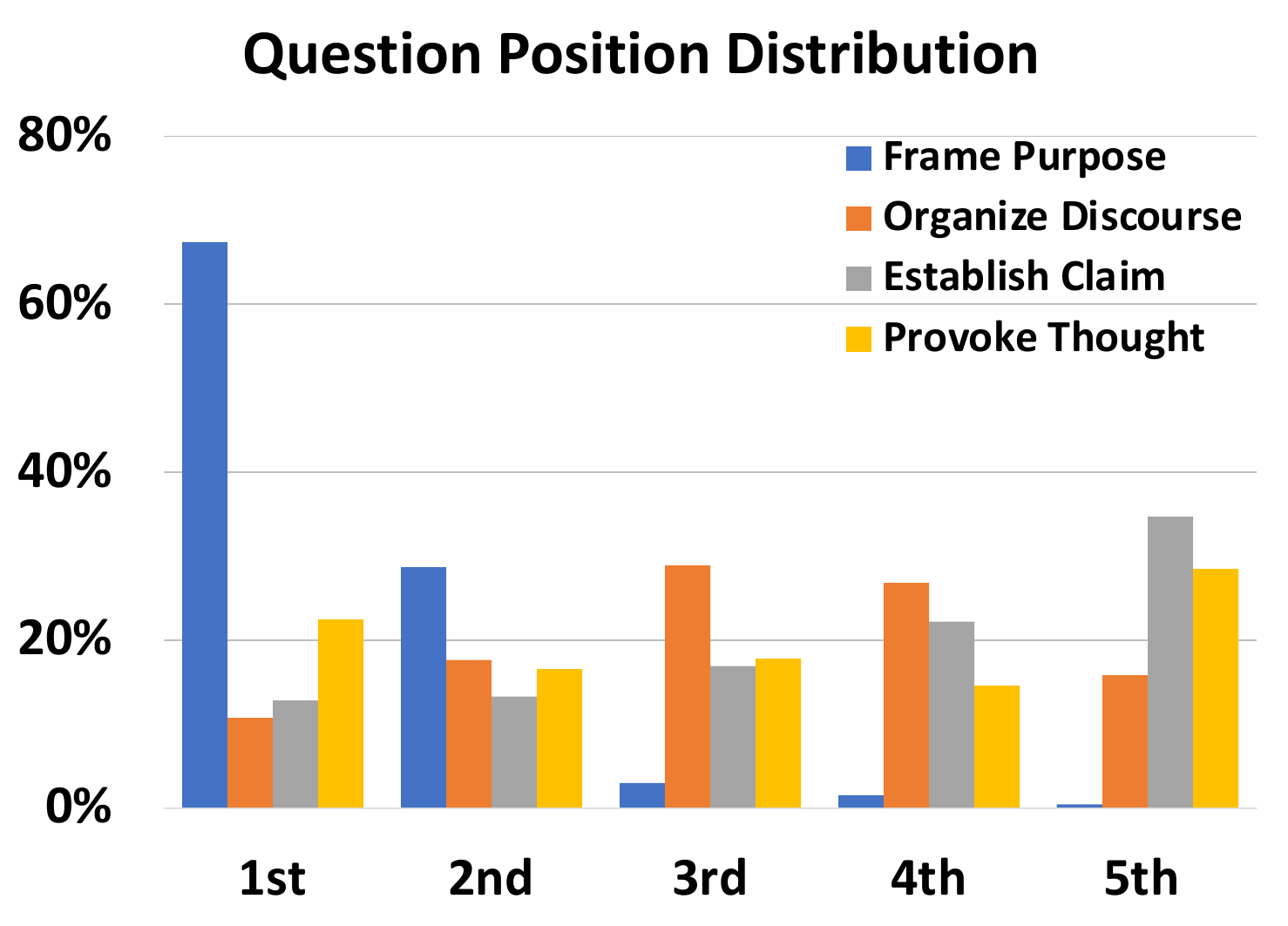}
    \caption{Position distribution of different questions. We omit \shading{Arouse Interest} questions as they are in titles by definition.}
    \label{fig:pos_dist}
    \vspace*{-4mm}
\end{figure}

\noindent \textbf{Where are guiding questions asked?$^{\spadesuit}$}
We divide articles into five equally sized segments and calculate the percentage of questions that appear in each part.
As in Figure \ref{fig:pos_dist}, there is a strong positional bias among different questions.
\shading{Frame Purpose} questions mostly appear in the first and second segments.
\shading{Organize Discourse} questions are largely located in the middle segments, while \shading{Provoke Thought} questions tend to emerge in the last segment, consistent with their expected functions.
\shading{Establish Claim} questions are skewed towards the end of the text. 
A possible explanation is that writers make more and more conclusive arguments with the progression of discussion.

\paragraph{Where are guiding questions answered?$^{\spadesuit}$}
We measure the distance between a question and its farthest evidence sentence (if any) in Figure 5. 
This can be considered the scope a question acts on the discourse of the document, i.e., from when the question is raised to when it is closed.
In this sense, \shading{Provoke Thought} questions go beyond the content as they do not have specific answers.
\begin{figure}[t]
    \centering
    \includegraphics[width=\columnwidth]{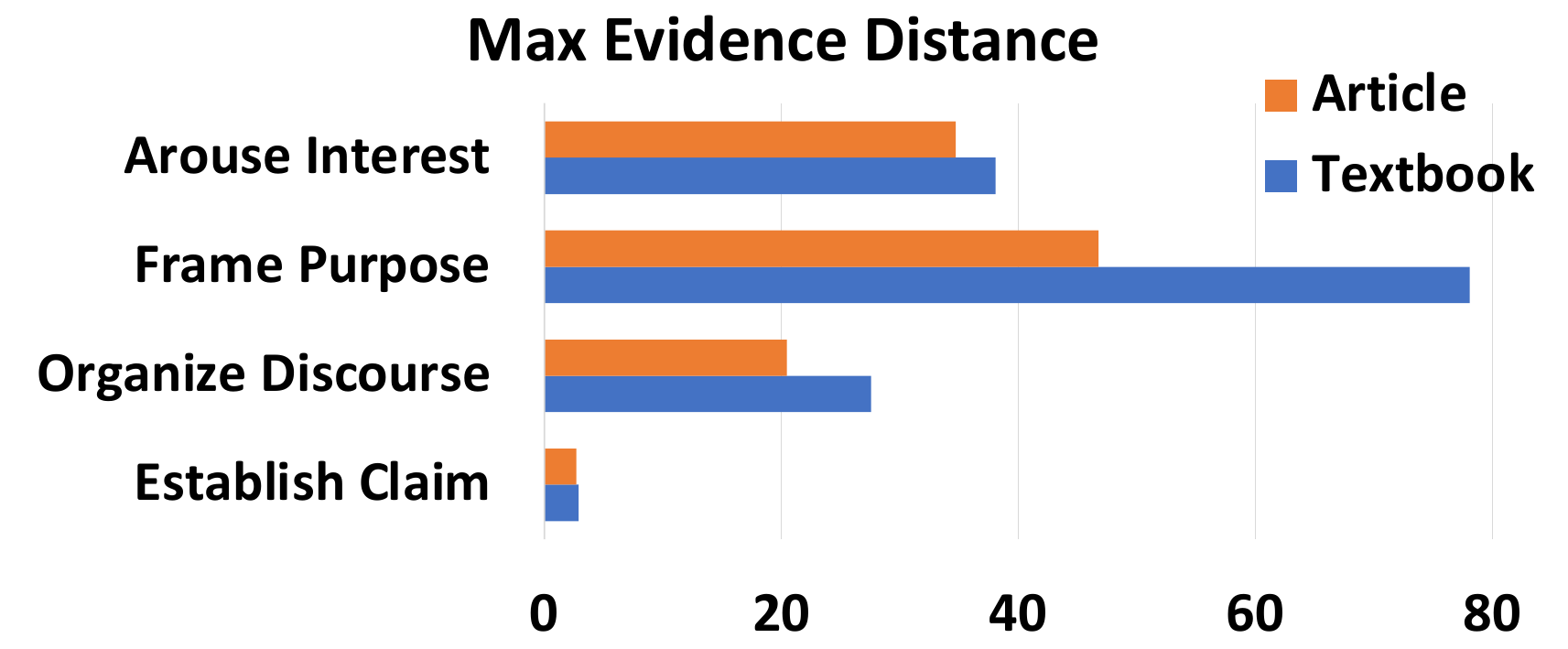}
    \caption{The average distance (in terms of sentence numbers) between a question and its farthest evidence.}
    \label{fig:evidence_dist}
    \vspace*{-4mm}
\end{figure}
We find the two subsets show a similar distribution. \shading{Frame Purpose} questions serve as the outline of a document, therefore having the largest range.
\shading{Organize Discourse} questions are usually answered within one or two paragraphs, consistent with their subheading role.
In contrast, \shading{Establish Claim} questions have prompt answers as the claims. 

\begin{figure}[h]
    \centering
    \includegraphics[width=\columnwidth]{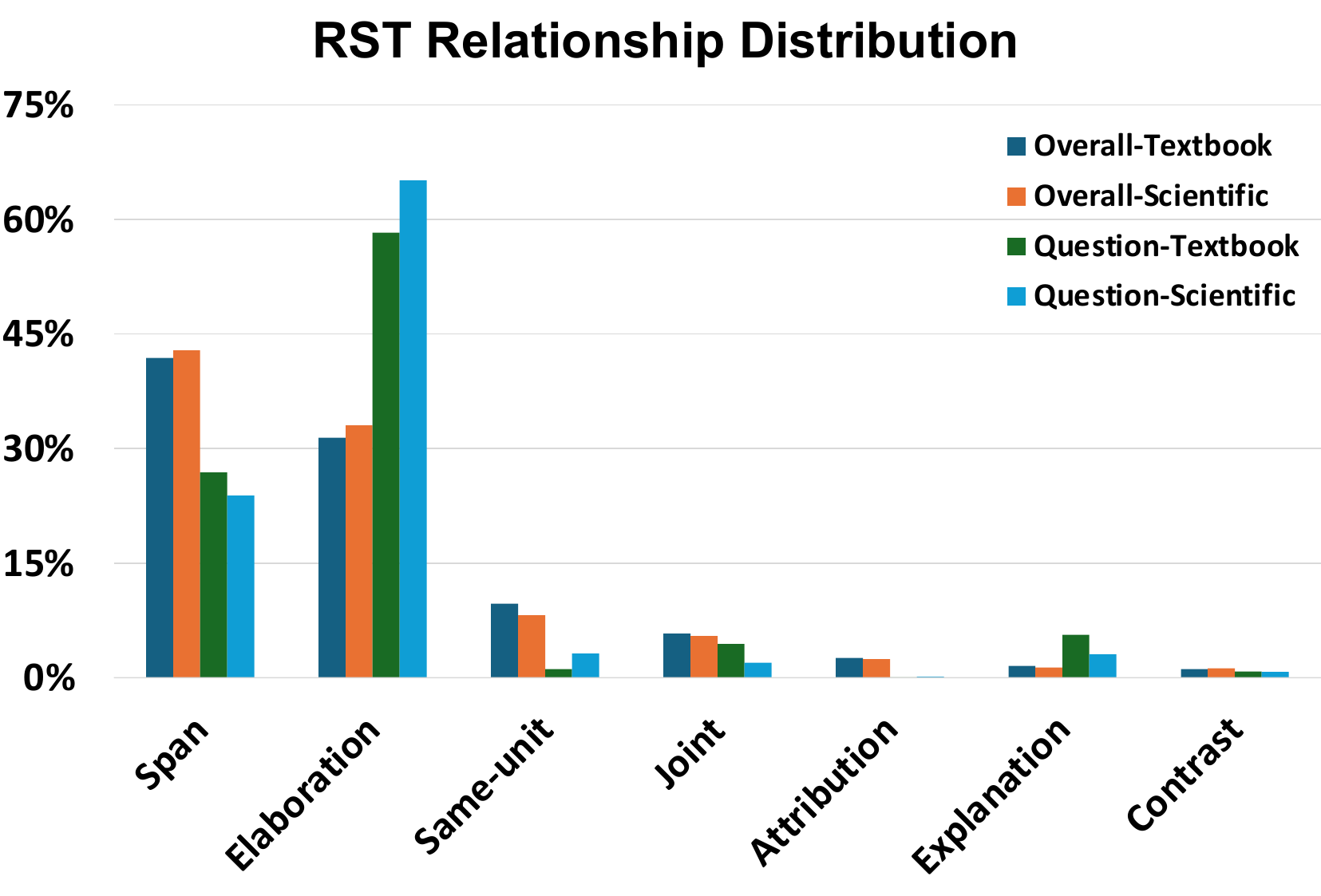}
    \caption{Distribution of RST discourse relationships with respect to questions and overall units. Relationships accounting for less than 1\% are omitted.}
    \label{fig:rst-distribution}
    \vspace*{-4mm}
\end{figure}
\paragraph{How are questions related to other text units?$^{\spadesuit}$}

As guiding questions are integral to the article, we also analyze the discourse relationships between questions and other text units\footnote{Following RST, we consider Elementary Discourse Unit (EDU) as the minimal unit.} within the same document using an RST \cite{mann1988rhetorical} parser\footnote{\href{https://github.com/EducationalTestingService/rstfinder}{github.com/EducationalTestingService/rstfinder}}.
In Figure \ref{fig:rst-distribution}, we compare the distribution of discourse relationships concerning questions with those of all text units. 
The analysis reveals that questions exhibit a higher proportion of concrete relationships, such as "elaboration" and "explanation," and a lower proportion of general relationships, such as "span" and "same-unit" compared to other units.
This suggests incorporating questions and their answers in writing might add to the coherence of the discourse structure.

\section{Guiding Question Generation}
\label{sec:QG}
In this section, we first describe our methods to model guiding questions and then report experimental results.
\subsection{Task Formulation}
Given a document $\mathcal{D}=\{s_{1}, ..., s_{n}\}$ of $n$ sentences, our goal is to learn a sequence of questions $\mathcal{Q}=\{q_1, ..., q_m\}$, their positions in the article $\mathcal{P}=\{p_1, ..., p_{m}\}$, and (optionally) their answer information $\mathcal{A}=\{a_1, ...,  a_{m}\}$.
In particular, $1 \leq p_{i} \leq n$ is the index of the sentence after which $q_{i}$ should be asked. 
We call these sentences \textbf{anchor sentences} $\{s_{p_{1}},...,s_{p_{m}}\}$.
\subsection{Data Preparation}
To construct training examples, we remove questions from articles and reconstruct them based on the corrupted articles such that the model can learn how to use guiding questions as human writers.
In concrete, given an article $\mathcal{D}$, we extract its questions and locate their positions to obtain $\mathcal{Q}$ and $\mathcal{P}$. 
For each question $q_{i}$, its answer information is a set of keywords $a_{i}=\{w_1, ...,w_{|a_{i}|}\}$ extracted from its evidence sentences obtained during the annotation (Section \ref{data-annotation}). 
We take this form to exclude redundant information in the full answer and reduce input (output) length for efficiency consideration.

Since deleting sentences would create incoherence and reduce learning $\mathcal{P}$ to identify where sentences are removed, we use \texttt{gpt-3.5-turbo-1106} to assess the coherence of missing positions and, if necessary, eliminate the incoherence by making small edits around the missing positions.
Since there could still be nuanced differences in edited positions, we perform the same ``delete and smooth'' operation on 1\% randomly selected non-question sentences as noise, detailed in Appendix \ref{subsec:del_smo}.

\subsection{Modeling}
In what follows, we describe three popular QG paradigms considered for our task: \textbf{Pipeline}, \textbf{Multitask}, and \textbf{Joint Generation} \cite{ushio-etal-2023-empirical}.
All approaches are unified as a text generation task and use Flan-T5 \cite{chung2022scaling} as the backbone model, which we finetune on our dataset.

\paragraph{Pipeline.} We decompose the task into three subtasks to learn $\{\mathcal{P}, \mathcal{A}, \mathcal{Q}\}$ with independent models. 

First, a \textbf{Position Predictor} (\textsc{PP}) identifies the positions of questions.
A naive way is to directly generate the position indices $\{p_{1}, ..., p_{m}\}$ conditioned on $\mathcal{D}$.
However, this requires learning a mapping between sentences and numerical symbols.
Instead, we opt to identify the \textit{anchor sentences} by training \textsc{PP} to copy them from $\mathcal{D}$:
\begin{gather}
        \tilde{C} = \mathop{{\rm argmax}}\limits_{C}P_{\theta_{\texttt{PP}}}(C|\mathcal{D}).
\end{gather} 
where the output $\tilde{C}=[s_{p_1}|...| s_{p_m}]$ is a concatenation of anchor sentences separated by "$|$".
$\mathcal{P}$ can be obtained by relocating copied sentences in $\mathcal{D}$. 
When there is no exact match, we use BM25 to get the most similar sentence.

Then, we highlight the target position in $\mathcal{D}$ by inserting a special mark ${\rm [Question]}$ after the anchor sentence and use an \textbf{Answer Extractor} (\textsc{AP}) to generate answer keywords $a_{i}$.:
\begin{gather}
        \mathcal{D}^{(i)} = [s_1, ..., s_{p_i}, {\rm[Question]},..., s_n], \\
        \tilde{a}_i = \mathop{{\rm argmax}}\limits_{a}P_{\theta_{\texttt{AE}}}(a|\mathcal{D}^{(i)}),
\end{gather}
where $\mathcal{D}^{(i)}$ is the document marked at position $p_{i}$. 

Finally, a \textbf{Question Generator} (\textsc{QG}) generates questions based on predicted positions and extracted answers:
\begin{gather}
    \tilde{q}_i = \mathop{{\rm argmax}}\limits_{q}P_{\theta_{\texttt{QG}}}(q| \mathcal{D}^{(i)}, a_{i}).
\end{gather}
Note that the \textsc{PP} model predicts all positions in one pass while the other two generate output one by one.
\paragraph{MultiTask.} The multitask model still consists of the three components described above. 
However, instead of independently training three models, we train a unified model for all tasks in a multitask learning manner.
In practice, we mix the training examples of the three tasks and distinguish them by adding different task prefixes before the inputs. 

\paragraph{Joint Model.} 
As observed in Figure \ref{fig:intro}, guiding questions tend to be related to each other.
Therefore, we consider jointly generating all questions at the same time.
To be specific, we use a template function to convert $\{\mathcal{P}, \mathcal{A}, \mathcal{Q}\}$ into a flattened sequence $\mathcal{T}(\mathcal{P}, \mathcal{A}, \mathcal{Q})=\{t(p_1,a_1, q_1)|t(p_2,a_2,q_2)...\}$ where $t(p, a, q) =$ "$\texttt{Position:} s_p \# \texttt{Answer:} a \# \texttt{Question:} q$". 
The model is trained to directly generate the whole target sequence $\mathcal{T}$ based on the article $\mathcal{D}$:
\begin{gather}
    \tilde{\mathcal{T}} = \mathop{{\rm argmax}}\limits_{\mathcal{T}}P_{\theta_{\texttt{JT}}}(\mathcal{T}| \mathcal{D}).
\end{gather}
In doing this, each question is also conditioned on previous ones, enabling the model to learn the inter-question connections.  

For the joint model, we also introduce a variant that additionally generates question roles.
This is done by inserting "\texttt{Role:} question role" between the position and answer field of each entry.
We denote this model as Joint$^{{\rm R}}$.

Since the above methods are all formalized as a text generation task, their training objectives take a similar form:
\begin{gather}
    \mathcal{L} = -\sum_{i}^{|Y|} \log P_{\theta}(y_{i}|y_{1}, ..., y_{i-1}, X),
\end{gather}
where $X$ and $Y$ are the input and output sequence of the corresponding task.
See Appendix \ref{subsec:training_setup} for training details.

\begin{table*}[htbp]
\centering
\small
\setlength\tabcolsep{5pt}
\begin{tabular}{@{}lccccc|ccccc@{}}
\toprule
\multirow{2}{*}{Models} & \multicolumn{5}{c}{\textbf{Textbook}}                                                                             & \multicolumn{5}{c}{\textbf{Scientific}}                                                                        \\ \cmidrule(l){2-11} 
                        & \textbf{\# Q}            & \textbf{Rouge-L} & \textbf{Meteor} & \textbf{BertScore} & \textbf{Dist-1/2} ($\downarrow$)            & \textbf{\# Q}           & \textbf{R-L}  & \textbf{Meteor} & \textbf{BertScore} & \textbf{Dist-1/2} ($\downarrow$)            \\ \midrule
GPT-4 (0-shot)          & 7.58                     & 15.7             & 18.9            & 84.3      & 64.4/48.9                     & 5.57                    & \textbf{14.3} & \textbf{19.7}   & 82.8               & 68.4/49.1                     \\ \midrule
Pipeline$_{\rm 250M}$                & 1.73                     & 12.9             & 13.7            & 77.4               & 71.4/51.6                     & 1.69                    & 12.7          & 7.93            & 79.5               & 66.1/48.3                     \\
Multitask$_{\rm 250M}$               & 1.65                     & 15.4             & 15.2            & 78.1               & 70.6/49.0                     & 1.84                    & 13.7          & 9.26            &  80.8             & 58.9/43.8                     \\
Joint$_{\rm 250M}$              & 3.41                     & 16.9             & 19.3            & 82.8               & 66.8/47.5                     & 2.08                    & 12.2          & 9.89            &  81.0              & 78.9/51.3                     \\
Joint$_{\rm 250M}^{{\rm R}}$          & 3.77                     & 18.3    & 20.8   & 84.5      & 65.4/46.9                     & 2.16                    & 13.1          & 10.2            & 81.0      & 75.2/49.8                     \\ \midrule
Joint$_{\rm 780M}^{{\rm R}}$          & 3.81                     & 30.5    & 28.1   & 87.7      & 65.6/47.1                     & 2.40                    & 12.4          & 9.38            & 82.5      & 71.3/48.9                     \\
Joint$_{\rm 3B}^{{\rm R}}$          & 3.95                     & 36.7    & 34.8   & 88.7      & 65.6/46.8                     & 2.46                    & 11.7          & 9.59            & 82.6      & 69.1/46.9                     \\
Joint$_{\rm 11B}^{{\rm R}}$          & 3.95                     & \textbf{56.4}    & \textbf{49.5}   & \textbf{92.1}      & 64.1/46.5                     & 2.55                    & 13.5          & 10.3            & \textbf{82.9}      & 68.5/47.2                     \\ \midrule
\rowcolor{lightgray} Reference               & 5.46 & 100.              & 100.             & 100.                & 61.7/46.3 & 4.50 & 100.           & 100.             & 100.                &66.7/48.9 \\ \bottomrule
\end{tabular}
\caption{Results of finetuned and prompt-based models on the \textsc{GuidingQ} dataset. Main takeaways: 1) human guiding questions are interrelated (as per Dist-N); 2) Joint generation w/ question role performs the best among fine-tuned models; 3)  Generated questions can capture the main message of human questions (as per BertScore).}
\label{tab:main_result}
\vspace{-4mm}
\end{table*}

\subsection{Automatic Evaluation}
The main results are presented in Table \ref{tab:main_result}, where we also include zero-shot prompted GPT-4 (\texttt{gpt-4-1104-preview}) with the prompt in Table \ref{tab:gpt4_prompt}.
We report the average number of generated questions \textbf{\# Q}, \textbf{Rouge (L)} \cite{lin-2004-rouge}, \textbf{Meteor} \cite{banerjee-lavie-2005-meteor}, and \textbf{BertScore} \cite{Zhang2020BERTScore:}.
Besides, we use \textbf{Dist-N (1/2)} \cite{li-etal-2016-diversity}, the percentage of distinct n-grams in generated questions, to measure the balance between \emph{diversity} and \emph{relevance} of guiding questions. 
Since there is no one-to-one map between generated and ground-truth questions, we concatenate all the questions as a whole sequence to compute reference-based metrics.

Overall, fine-tuned models generate fewer questions than references, while zero-shot GPT-4 generates more.   
We found that Flan-T5 tends to output short sequences; therefore, we attribute this to their different pre-training paradigms. 
Jointly generating question roles can boost performance, which is expected as they are indicative of a series of distributional features. 
The best fine-tuned models achieve remarkable BertScores.
This suggests that the generated questions successfully replicate the main information of human questions.

For the textbook dataset, the joint model generally performs the best among fine-tuned approaches.
In particular, it best resembles the Dist-1/2 results of human questions, while others tend to generate more independent questions (higher Dist-1/2).
This suggests that the success of this model is possibly because it better learns the inter-question relationships. 
As for the scientific set, the multitask model achieves competitive results with the joint model.
We conjecture the more complex content and sparser questions increase the difficulty of learning the question relationship, which could diminish the advantage of joint generation.

Finally, we scale up the parameter size of the best-performing model Joint$^{{\rm R}}$ up to 11B. 
We can see that scaling the model size results in significant performance gain on the textbook set but little on the scientific set.
This demonstrates the challenge of understanding scientific language with LLMs.

\paragraph{Question Position.} 
Since the generated questions are different from references, naively matching their positions (e.g., recall, precision) is not a suitable way. 
Instead, we evaluate a question's position by measuring how well it fits the context using perplexity. 
Specifically, let $p$ be the index of question $q$. 
We insert $q$ in $p$ and calculate the average word perplexity of sentences $[s_{p-3}, s_{p-2}, s_{p-1}, q, s_{p+1}, s_{p+2}, s_{p+3}]$, each conditioned on its previous three sentences. 
\begin{figure}[t]
    \centering
    \includegraphics[width=\linewidth]{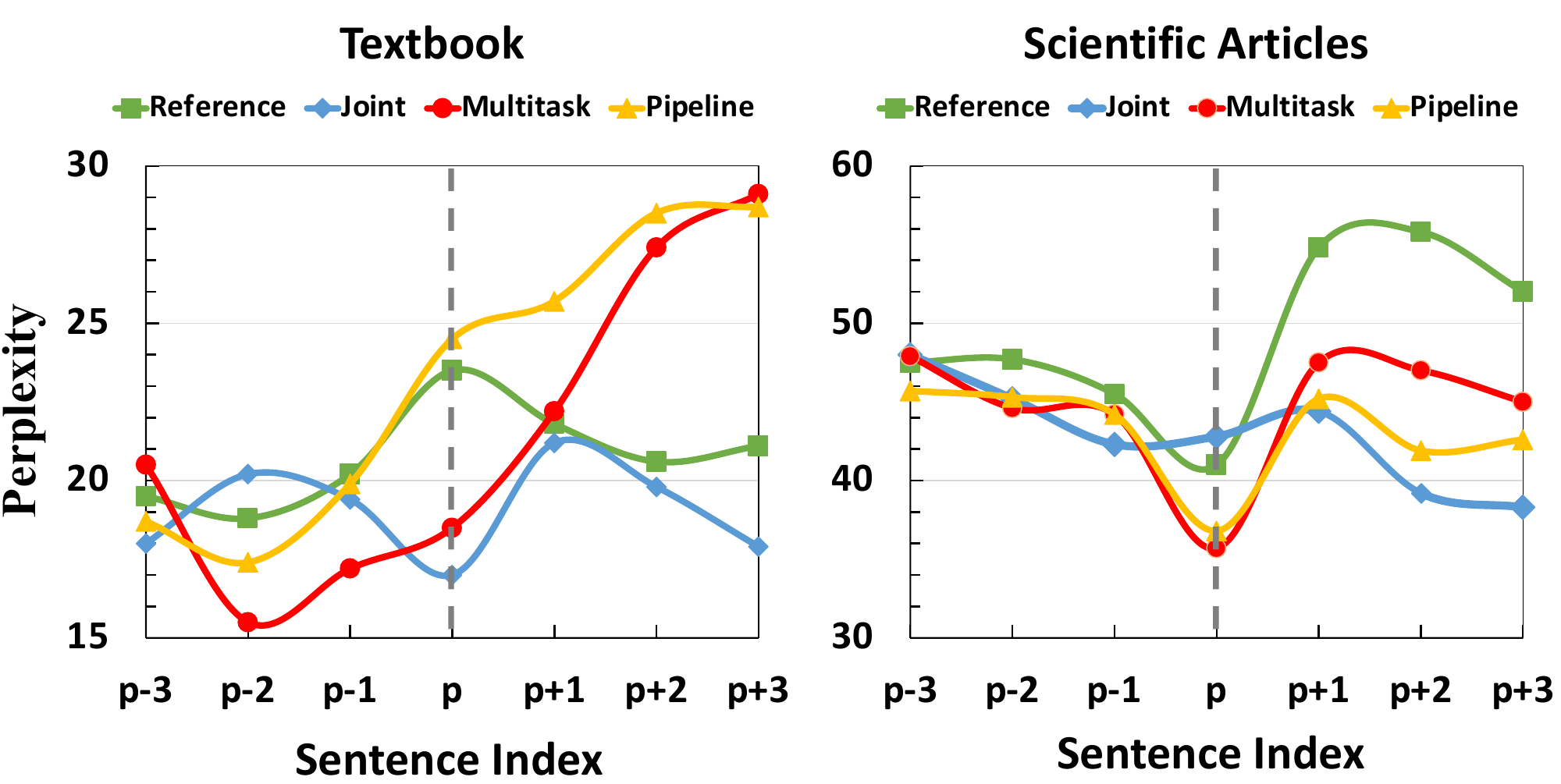}
    \vspace*{-4mm}
    \caption{Average word perplexity of the question (index $p$) and its surrounding sentences. We omit questions by 0-shot GPT-4 as we find their positions are sensitive to prompts and temperatures.}
    \label{fig:position_ppl}
    \vspace*{-5mm}
\end{figure}
The results are presented in Figure \ref{fig:position_ppl}. 
Interestingly, human questions in textbooks create a local peak of PPL in the context. 
We explain this phenomenon from the connection between \emph{surprisal} and \emph{salience}: text units with higher salience are usually less expected, making them stand out from the context and attract the reader’s attention \cite{racz2013salience,zarcone2016salience,blumenthal2017perceptual}. 
This is consistent with the interactional purpose of using questions. 
However, none of the models replicate this salience effect, either lowering the PPL (joint model) or increasing the PPL without a quick decline (pipeline/multitask).

The scientific articles show a different pattern where human questions are with lower PPL. 
We conjecture this is because scientific articles usually make sufficient discussions before proposing a question. 
As a consequence, their questions are highly predictable by prior sentences. 
Finetuned models manage to replicate the effect, possibly because research articles and their question usage are usually structured (e.g., proposing questions in the introduction) and thus easier to learn. 

\section{Human Study} \label{Section-Human-Study}
Finally, we conduct a between-group human study to investigate the impact of guiding questions on reading comprehension.
Participants are asked to read articles with (or without) questions, after which we gather their feedback and analyze their information retention and understanding to gain a holistic view of the effect of guiding questions.

\begin{table}[t]
\centering
\begin{tabular}{l<{\hspace{-2mm}}lcc}
\toprule
 && \bf Reference & \bf Generated \\
\midrule
\parbox[t]{2mm}{\multirow{3}{*}{\rotatebox[origin=c]{90}{\small \bf Quality}}}

& Relevance &
4.0 \includegraphics[height=4mm]{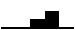} &
4.2 \includegraphics[height=4mm]{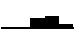} \\

& Position &
4.2 \includegraphics[height=4mm]{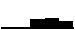} & 
4.3 \includegraphics[height=4mm]{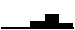} \\

& Importance &
4.4 \includegraphics[height=4mm]{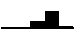} &
4.2 \includegraphics[height=4mm]{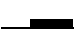} \\

\cmidrule{1-2}

\parbox[t]{2mm}{\multirow{3}{*}{\rotatebox[origin=c]{90}{\small \bf Usefulness}}}
& Engaging &
3.9 \includegraphics[height=4mm]{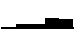} &
4.0 \includegraphics[height=4mm]{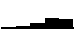} \\

& Understanding &
4.2 \includegraphics[height=4mm]{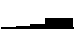} &
4.1 \includegraphics[height=4mm]{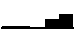} \\

& Overall & 
4.3 \includegraphics[height=4mm]{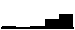} &
3.9 \includegraphics[height=4mm]{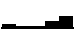} \\

\bottomrule
\end{tabular}
\caption{Average scores by participants. ``Quality'' is evaluated on each single question, while ``Usefulness'' is evaluated on all questions of an article.}
\label{tab:human-quantitative}
\vspace{-4mm}
\end{table}

\subsection{Experiment Design}
\label{subsec: exp_design}
\textbf{Procedure.} The human study runs on a crowdsourcing platform \href{https://www.prolific.com/}{Prolific} and consists of 4 stages.
\begin{enumerate}[left=0mm,noitemsep,topsep=1mm]
\item \textbf{Demographic Questionnaire}: We collect participants' demographic information (Appendix \ref{subsec:demo_info}) and consent before the experiment.
\item \textbf{Article Reading}: Participants read an assigned article in 20 minutes with or without questions beside the article. Details of the reading interface are in Appendix \ref{sub:app_read_interface}.  
\item \textbf{Comprephension Test}: After reading, participants are asked to write a summary of at least 100 words \emph{without} access to the article.
\item \textbf{Evaluation}: Finally, we ask participants to rate the usefulness and quality of questions (If any).
\end{enumerate}

\vspace{1.5mm}
\noindent \textbf{Participants.} We select 45 participants with the criteria being at least C1 command of English.
Each participant received 10 GBP/hour on average.
We randomly assign them to one of three groups:
\begin{itemize}[left=0mm,noitemsep,topsep=1mm]
\item \textbf{Control}: reading w/o questions.
\item \textbf{Reference}: reading w/ expert-written questions.
\item \textbf{Generated}: reading w/ generated questions.
\end{itemize}

\vspace{1.5mm}
\noindent \textbf{Test Articles.} We use the Joint$^{\textsc{R}}$ model to generate questions for five textbook chapters\footnote{We only consider textbook passages because scientific articles are challenging for general readers.} selected from different domains spanning \emph{Business}, \emph{Philosophy}, \emph{Sociology}, \emph{Political Science}, and \emph{Psychology}.
A few small adaptations are made to make them better fit the study (e.g., length reduction).
To focus on question quality, we generate the same number of questions as reference\footnote{We do this by truncating over-generated questions or disallowing the \textless{}eos\textgreater{} token until we get enough questions.}. 
Each article is assigned to 3 participants to average-out the effect of individual articles. 

\subsection{Question Evaluation}
Participants evaluate the intrinsic quality of each question from three aspects: \emph{relevance}, \emph{position}, and \emph{importance} (detailed in Table \ref{tab:human_eval_questions}).
From a later survey, we found that participants may have different interpretations of these dimensions despite our instructions. 
Nevertheless, the results in the first section of Table \ref{tab:human-quantitative} prove that the average quality of generated questions is on par with (if not better than) ground truth ones.

\begin{table}[t]
\small
\centering
\begin{tabular}{@{}lccc@{}}
\toprule
\multirow{2}{*}{} & \multicolumn{3}{c}{\textbf{Summary Quality (1-5)}}                   \\ \cmidrule(l){2-4} 
                  & \textbf{Coherence} & \textbf{Consistency} & \textbf{Informativeness} \\ \midrule
\textbf{Ctrl.}    & 2.38$_{\pm0.09}$              & 2.52$_{\pm0.08}$                & 2.31$_{\pm0.09}$                  \\
\textbf{Gen.}     & 3.14$_{\pm0.08}$               & \textbf{3.28$_{\pm0.08}$}        & 2.98$_{\pm0.05}$                     \\
\textbf{Ref.}    & \textbf{3.28$_{\pm0.04}$}      & 3.23$_{\pm0.06}$                 & \textbf{2.99}$_{\pm0.08}$            \\ \bottomrule
\end{tabular}
\caption{Quality scores (Mean$_{{\rm std.}}$) of summaries from three groups over 5 runs. }
\label{tab:summary_eval}
\vspace{-4mm}
\end{table}
\subsection{Effect on Reading Comprehension}
\paragraph{User Perceived Usefulness.}
Results in the second part of Table \ref{tab:human-quantitative} prove that the generated questions are as good as human questions in perceived usefulness. 
See the questionnaire in Table \ref{tab:human_eval_questions}.
\vspace{1.5mm}

\noindent \textbf{Improved Memorization and Comprehension.}
We use summary quality as a proxy to measure participants' memorization and comprehension.
The evaluation consists of three dimensions: \emph{Coherence}, \emph{Consistency}, and \emph{Informativeness}, and is based on GPT-4, which has shown a superior correlation with human annotations on summary evaluation \cite{liu-etal-2023-g}, detailed in Appendix \ref{sec:app_sum_eval}.
As shown in Table \ref{tab:summary_eval}, the reference and generated groups achieve comparable scores, and both outperform the control group by a large margin.
An additional between-group summary analysis, including summary time, length, and n-gram overlap, is shown in Table \ref{tab:add_summary}. 


An intuitive explanation for the improved quality is that users memorize question-related information and incorporate them into the summary.
To verify this, we measured the \emph{entailment} score between summaries and answers to guiding questions using BertScore recall (${\rm BS}_r$):
\begin{gather}
    \textsc{EntScore} = \frac{1}{|S|}\sum_{s \in S}\frac{1}{|Q_{s}|}\sum_{q \in Q_{s}}{\rm BS}_{r}(s, a_q),
\end{gather}
where $s$ is a summary, $Q_{s}$ is the set of guiding questions of the source article of $s$, and $a_q$ is the answer to $q$.

The results in Table \ref{tab:EntScore} show that the reference and generated groups incorporated more answer information compared to the control group, indicating improved memorization. 
However, the difference in recall scores is not as pronounced as the difference in summary quality (Table \ref{tab:summary_eval}). 
This suggests that readers do not simply compile question-answer pairs into summaries. 
Therefore, we believe the improvement in summary quality also reflects a deeper understanding facilitated by the guiding questions. 
\vspace{1.5mm}

\begin{table}[t]
\centering
\small
\begin{tabular}{@{}lcc@{}}
\toprule
\multicolumn{1}{l}{\diagbox{\textbf{Sum.}}{\textbf{Ans.}}} & \textbf{Reference} & \textbf{Generated} \\ \midrule
\textbf{Reference}             & \textbf{55.2}      & --                  \\
\textbf{Generated}             & --                  & \textbf{56.3}      \\ \midrule
\textbf{Control}               & 52.8               & 52.5               \\ \bottomrule
\end{tabular}
\caption{Entailment score between guiding questions summaries and answers with BertScore recall ($\times 100$).}
\label{tab:EntScore}
\vspace{-4mm}
\end{table}

\noindent \textbf{Reading Time.}
The reading speeds of different groups are shown in Table \ref{tab:reading_speed}.
When questions are displayed next to the articles, the users spend a longer time reading.
On the one hand, this is a signal of enhanced engagement.
On the other hand, this potentially implies an extra cognitive load caused by guiding questions.

\begin{table}[htbp]
\small
\centering
\begin{tabular}{@{}cccc@{}}
\toprule
                             & \textbf{Auth.} & \textbf{Gen.} & \textbf{Contr.} \\ \midrule
\textbf{Reading Speed (w/s)} & 1.00           & 0.98          & 1.39            \\ \bottomrule
\end{tabular}
\caption{Reading speeds of different user groups measured by words per second (w/s).}
\label{tab:reading_speed}
\vspace{-3mm}
\end{table}

\vspace{1mm}

\noindent \textbf{Participant Feedback.} 
Finally, we gathered user feedback through a preliminary interview-based study, detailed in Appendix \ref{Appendix: observation}. 
In summary, participants confirmed the benefits of guiding questions and discussed various aspects of these questions, highlighting both consistent preferences and nuanced complexities. 
We hope these insights will inspire future research in this area.

\section{Conclusion}
This paper studies the discourse and interactional role of guiding questions in textbooks and scientific articles. 
We explore various approaches for modeling these questions, providing insights into how to model this task and highlighting challenges to be solved.
We validate our results with human studies, which demonstrate reading with guiding questions can improve the high-level memorization and understanding of human readers.

\section*{Broader Impacts}
In this study, we analyzed the use of questions in academic and educational articles, demonstrating their benefits for reading comprehension. 
While questions can enhance engagement, they can also increase readers' cognitive load, as evidenced by longer reading times (Table \ref{tab:reading_speed}). 
Additionally, questions may introduce unintended nuances for communication, such as creating unequal social relationships \cite{hyland2002they}. 
Therefore, it is important to be aware of these mixed effects when using guiding questions in writing.

\section*{Limitations}
We summarize the limitations of this study into the following open questions.

\paragraph{Is our question role taxonomy generalizable to other domains?$^{\bigstar}$} Our investigation of the role of guiding questions is initially focused on textbooks and scientific articles. 
However, different domains might use questions differently. 
Nevertheless, our analysis (Section \ref{subsec:data_analysis}) uncovers distributional features that are indicative of question functions, such as their positions and question-answer relationships. These findings offer insights that could be generalized to understand the roles of questions in broader contexts.

\paragraph{How to align guiding questions with individual preferences?$^{\bigstar}$} Our model aims to replicate guiding questions crafted by human writers. 
However, these questions may not always resonate with individual readers, given their different reading goals and prior knowledge. 
We expect that personalized generation \cite{cui-sachan-2023-adaptive}, which takes into account user profiles, would yield more helpful questions.

\paragraph{How has the role of questions evolved?$^{\bigstar}$} 

It is important to note that the use of questions could change over time. 
For instance, \citet{ball2009scholarly,jiang2022titles} have analyzed the distribution shift of questions in titles over the past decades. 
In this study, we did not take the temporal dimension into account, and the conclusions are based on contemporary texts.
Therefore, the findings of this paper may not remain consistent in the future.


\vspace{3mm}

\begin{thebibliography}{39}
\expandafter\ifx\csname natexlab\endcsname\relax\def\natexlab#1{#1}\fi

\bibitem[{Ball(2009)}]{ball2009scholarly}
Rafael Ball. 2009.
\newblock Scholarly communication in transition: The use of question marks in the titles of scientific articles in medicine, life sciences and physics 1966--2005.
\newblock \emph{Scientometrics}, 79(3):667--679.

\bibitem[{Banerjee and Lavie(2005)}]{banerjee-lavie-2005-meteor}
Satanjeev Banerjee and Alon Lavie. 2005.
\newblock \href {https://aclanthology.org/W05-0909} {{METEOR}: An automatic metric for {MT} evaluation with improved correlation with human judgments}.
\newblock In \emph{Proceedings of the {ACL} Workshop on Intrinsic and Extrinsic Evaluation Measures for Machine Translation and/or Summarization}, pages 65--72, Ann Arbor, Michigan. Association for Computational Linguistics.

\bibitem[{Benz and Jasinskaja(2017)}]{benz2017questions}
Anton Benz and Katja Jasinskaja. 2017.
\newblock \href {https://www.tandfonline.com/doi/full/10.1080/0163853X.2017.1316038} {Questions under discussion: {From} sentence to discourse}.

\bibitem[{Bharuthram(2017)}]{bharuthram2017facilitating}
Sharita Bharuthram. 2017.
\newblock \href {https://www.ajol.info/index.php/jlt/article/view/180139} {Facilitating active reading through a self-questioning strategy: {Student} and tutor experiences and reflections of the strategy use}.
\newblock \emph{Journal for Language Teaching}, 51(2):85--103.

\bibitem[{Blumenthal-Dram{\'e} et~al.(2017)Blumenthal-Dram{\'e}, Hanul{\'\i}kov{\'a}, and Kortmann}]{blumenthal2017perceptual}
Alice Blumenthal-Dram{\'e}, Adriana Hanul{\'\i}kov{\'a}, and Bernd Kortmann. 2017.
\newblock \href {https://www.frontiersin.org/journals/psychology/articles/10.3389/fpsyg.2017.00411/full} {Perceptual linguistic salience: {Modeling} causes and consequences}.

\bibitem[{Chung et~al.(2022)Chung, Hou, Longpre, Zoph, Tay, Fedus, Li, Wang, Dehghani, Brahma et~al.}]{chung2022scaling}
Hyung~Won Chung, Le~Hou, Shayne Longpre, Barret Zoph, Yi~Tay, William Fedus, Yunxuan Li, Xuezhi Wang, Mostafa Dehghani, Siddhartha Brahma, et~al. 2022.
\newblock \href {https://www.jmlr.org/papers/v25/23-0870.html} {Scaling instruction-finetuned language models}.
\newblock \emph{arXiv preprint arXiv:2210.11416}.

\bibitem[{Cohan et~al.(2018)Cohan, Dernoncourt, Kim, Bui, Kim, Chang, and Goharian}]{cohan-etal-2018-discourse}
Arman Cohan, Franck Dernoncourt, Doo~Soon Kim, Trung Bui, Seokhwan Kim, Walter Chang, and Nazli Goharian. 2018.
\newblock \href {https://doi.org/10.18653/v1/N18-2097} {A discourse-aware attention model for abstractive summarization of long documents}.
\newblock In \emph{Proceedings of the 2018 Conference of the North {A}merican Chapter of the Association for Computational Linguistics: Human Language Technologies, Volume 2 (Short Papers)}, pages 615--621, New Orleans, Louisiana. Association for Computational Linguistics.

\bibitem[{Cui and Sachan(2023)}]{cui-sachan-2023-adaptive}
Peng Cui and Mrinmaya Sachan. 2023.
\newblock \href {https://doi.org/10.18653/v1/2023.acl-long.567} {Adaptive and personalized exercise generation for online language learning}.
\newblock In \emph{Proceedings of the 61st Annual Meeting of the Association for Computational Linguistics (Volume 1: Long Papers)}, pages 10184--10198, Toronto, Canada. Association for Computational Linguistics.

\bibitem[{Curry and Chambers(2017)}]{curry2017questions}
Niall Curry and Angela Chambers. 2017.
\newblock \href {https://link.springer.com/article/10.1007/s41701-017-0012-0} {Questions in english and french research articles in linguistics: {A} corpus-based contrastive analysis}.
\newblock \emph{Corpus Pragmatics}, 1:327--350.

\bibitem[{De~Kuthy et~al.(2018)De~Kuthy, Reiter, and Riester}]{de-kuthy-etal-2018-qud}
Kordula De~Kuthy, Nils Reiter, and Arndt Riester. 2018.
\newblock \href {https://aclanthology.org/L18-1304} {{QUD}-based annotation of discourse structure and information structure: Tool and evaluation}.
\newblock In \emph{Proceedings of the Eleventh International Conference on Language Resources and Evaluation ({LREC} 2018)}, Miyazaki, Japan. European Language Resources Association (ELRA).

\bibitem[{Haggan(2004)}]{haggan2004293}
Madeline Haggan. 2004.
\newblock \href {https://doi.org/https://doi.org/10.1016/S0378-2166(03)00090-0} {Research paper titles in literature, linguistics and science: {Dimensions} of attraction}.
\newblock \emph{Journal of Pragmatics}, 36(2):293--317.

\bibitem[{He et~al.(2023)He, Lin, Gong, Jin, Zhang, Lin, Jiao, Yiu, Duan, Chen et~al.}]{he2023annollm}
Xingwei He, Zhenghao Lin, Yeyun Gong, Alex Jin, Hang Zhang, Chen Lin, Jian Jiao, Siu~Ming Yiu, Nan Duan, Weizhu Chen, et~al. 2023.
\newblock \href {https://arxiv.org/abs/2303.16854} {Annollm: {Making} large language models to be better crowdsourced annotators}.
\newblock \emph{arXiv preprint arXiv:2303.16854}.

\bibitem[{Hyland(2002)}]{hyland2002they}
Ken Hyland. 2002.
\newblock \href {https://www.degruyter.com/document/doi/10.1515/text.2002.021/html} {What do they mean? questions in academic writing}.
\newblock \emph{Text \& Talk}, 22(4):529--557.

\bibitem[{Jamali and Nikzad(2011)}]{jamali2011article}
Hamid~R Jamali and Mahsa Nikzad. 2011.
\newblock \href {https://akjournals.com/view/journals/11192/88/2/article-p653.xml} {Article title type and its relation with the number of downloads and citations}.
\newblock \emph{Scientometrics}, 88(2):653--661.

\bibitem[{Jiang and Hyland(2022)}]{jiang2022titles}
Feng~Kevin Jiang and Ken Hyland. 2022.
\newblock \href {https://onlinelibrary.wiley.com/doi/full/10.1002/leap.1498} {Titles in research articles: {Changes} across time and discipline}.
\newblock \emph{Learned Publishing}.

\bibitem[{Ko et~al.(2022)Ko, Dalton, Simmons, Fisher, Durrett, and Li}]{ko-etal-2022-discourse}
Wei-Jen Ko, Cutter Dalton, Mark Simmons, Eliza Fisher, Greg Durrett, and Junyi~Jessy Li. 2022.
\newblock \href {https://doi.org/10.18653/v1/2022.emnlp-main.806} {Discourse comprehension: {A} question answering framework to represent sentence connections}.
\newblock In \emph{Proceedings of the 2022 Conference on Empirical Methods in Natural Language Processing}, pages 11752--11764, Abu Dhabi, United Arab Emirates. Association for Computational Linguistics.

\bibitem[{Li et~al.(2016)Li, Galley, Brockett, Gao, and Dolan}]{li-etal-2016-diversity}
Jiwei Li, Michel Galley, Chris Brockett, Jianfeng Gao, and Bill Dolan. 2016.
\newblock \href {https://doi.org/10.18653/v1/N16-1014} {A diversity-promoting objective function for neural conversation models}.
\newblock In \emph{Proceedings of the 2016 Conference of the North {A}merican Chapter of the Association for Computational Linguistics: Human Language Technologies}, pages 110--119, San Diego, California. Association for Computational Linguistics.

\bibitem[{Li et~al.(2023)Li, Shi, Ziems, Kan, Chen, Liu, and Yang}]{li-etal-2023-coannotating}
Minzhi Li, Taiwei Shi, Caleb Ziems, Min-Yen Kan, Nancy Chen, Zhengyuan Liu, and Diyi Yang. 2023.
\newblock \href {https://doi.org/10.18653/v1/2023.emnlp-main.92} {{C}o{A}nnotating: {Uncertainty-Guided} work allocation between human and large language models for data annotation}.
\newblock In \emph{Proceedings of the 2023 Conference on Empirical Methods in Natural Language Processing}, pages 1487--1505, Singapore. Association for Computational Linguistics.

\bibitem[{Lin(2004)}]{lin-2004-rouge}
Chin-Yew Lin. 2004.
\newblock \href {https://aclanthology.org/W04-1013} {{ROUGE}: A package for automatic evaluation of summaries}.
\newblock In \emph{Text Summarization Branches Out}, pages 74--81, Barcelona, Spain. Association for Computational Linguistics.

\bibitem[{Liu et~al.(2023)Liu, Iter, Xu, Wang, Xu, and Zhu}]{liu-etal-2023-g}
Yang Liu, Dan Iter, Yichong Xu, Shuohang Wang, Ruochen Xu, and Chenguang Zhu. 2023.
\newblock \href {https://doi.org/10.18653/v1/2023.emnlp-main.153} {{G}-eval: {NLG} evaluation using gpt-4 with better human alignment}.
\newblock In \emph{Proceedings of the 2023 Conference on Empirical Methods in Natural Language Processing}, pages 2511--2522, Singapore. Association for Computational Linguistics.

\bibitem[{Mann and Thompson(1988)}]{mann1988rhetorical}
William~C Mann and Sandra~A Thompson. 1988.
\newblock \href {https://www.degruyter.com/document/doi/10.1515/text.1.1988.8.3.243/html} {Rhetorical structure theory: {Toward} a functional theory of text organization}.
\newblock \emph{Text-interdisciplinary Journal for the Study of Discourse}, 8(3):243--281.

\bibitem[{Nallapati et~al.(2017)Nallapati, Zhai, and Zhou}]{nallapati2017summarunner}
Ramesh Nallapati, Feifei Zhai, and Bowen Zhou. 2017.
\newblock \href {https://ojs.aaai.org/index.php/AAAI/article/view/10958} {Summarunner: {A} recurrent neural network based sequence model for extractive summarization of documents}.
\newblock In \emph{Proceedings of the AAAI conference on artificial intelligence}, volume 31/1.

\bibitem[{R{\'a}cz(2013)}]{racz2013salience}
P{\'e}ter R{\'a}cz. 2013.
\newblock \href {https://www.jstor.org/stable/24770229} {\emph{Salience in sociolinguistics: {A} quantitative approach}}, volume~84.
\newblock Walter de Gruyter.

\bibitem[{Raffel et~al.(2020)Raffel, Shazeer, Roberts, Lee, Narang, Matena, Zhou, Li, and Liu}]{raffel2020exploring}
Colin Raffel, Noam Shazeer, Adam Roberts, Katherine Lee, Sharan Narang, Michael Matena, Yanqi Zhou, Wei Li, and Peter~J Liu. 2020.
\newblock \href {https://www.jmlr.org/papers/v21/20-074.html} {Exploring the limits of transfer learning with a unified text-to-text transformer}.
\newblock \emph{The Journal of Machine Learning Research}, 21(1):5485--5551.

\bibitem[{Roberts(2012)}]{roberts2012information}
Craige Roberts. 2012.
\newblock \href {https://semprag.org/index.php/sp/article/view/sp.5.6} {Information structure: {Towards} an integrated formal theory of pragmatics}.
\newblock \emph{Semantics and pragmatics}, 5:6--1.

\bibitem[{Shazeer and Stern(2018)}]{shazeer2018adafactor}
Noam Shazeer and Mitchell Stern. 2018.
\newblock \href {http://proceedings.mlr.press/v80/shazeer18a.html?ref=https://githubhelp.com} {Adafactor: {Adaptive} learning rates with sublinear memory cost}.
\newblock In \emph{International Conference on Machine Learning}, pages 4596--4604. PMLR.

\bibitem[{Singh et~al.(2023)Singh, Zouhar, and Sachan}]{singh-etal-2023-enhancing}
Janvijay Singh, Vil{\'e}m Zouhar, and Mrinmaya Sachan. 2023.
\newblock \href {https://aclanthology.org/2023.emnlp-main.731} {Enhancing textbooks with visuals from the web for improved learning}.
\newblock In \emph{Proceedings of the 2023 Conference on Empirical Methods in Natural Language Processing}, pages 11931--11944, Singapore. Association for Computational Linguistics.

\bibitem[{Syamsiah et~al.(2018)Syamsiah, Rafli, and Ridwan}]{syamsiah2018self}
Nur Syamsiah, Zainal Rafli, and Sakura Ridwan. 2018.
\newblock \href {https://www.atlantis-press.com/proceedings/aecon-18/55908989} {Self--questioning strategy on reading comprehension process}.
\newblock In \emph{5th Asia Pasific Education Conference (AECON 2018)}, pages 120--129. Atlantis Press.

\bibitem[{T{\"o}rnberg(2023)}]{tornberg2023chatgpt}
Petter T{\"o}rnberg. 2023.
\newblock \href {https://arxiv.org/abs/2304.06588} {Chatgpt-4 outperforms experts and crowd workers in annotating political twitter messages with zero-shot learning}.
\newblock \emph{arXiv preprint arXiv:2304.06588}.

\bibitem[{Ushio et~al.(2023)Ushio, Alva-Manchego, and Camacho-Collados}]{ushio-etal-2023-empirical}
Asahi Ushio, Fernando Alva-Manchego, and Jose Camacho-Collados. 2023.
\newblock \href {https://doi.org/10.18653/v1/2023.findings-acl.899} {An empirical comparison of {LM}-based question and answer generation methods}.
\newblock In \emph{Findings of the Association for Computational Linguistics: ACL 2023}, pages 14262--14272, Toronto, Canada. Association for Computational Linguistics.

\bibitem[{Van~Kuppevelt(1995)}]{van1995discourse}
Jan Van~Kuppevelt. 1995.
\newblock \href {https://www.cambridge.org/core/journals/journal-of-linguistics/article/discourse-structure-topicality-and-questioning/60F3E68601517091AF560CB3CC02C6AC} {Discourse structure, topicality and questioning}.
\newblock \emph{Journal of linguistics}, 31(1):109--147.

\bibitem[{Wei et~al.(2022)Wei, Wang, Schuurmans, Bosma, ichter, Xia, Chi, Le, and Zhou}]{NEURIPS2022_9d560961}
Jason Wei, Xuezhi Wang, Dale Schuurmans, Maarten Bosma, brian ichter, Fei Xia, Ed~Chi, Quoc~V Le, and Denny Zhou. 2022.
\newblock \href {https://proceedings.neurips.cc/paper_files/paper/2022/file/9d5609613524ecf4f15af0f7b31abca4-Paper-Conference.pdf} {Chain-of-thought prompting elicits reasoning in large language models}.
\newblock In \emph{Advances in Neural Information Processing Systems}, volume~35, pages 24824--24837. Curran Associates, Inc.

\bibitem[{Westera et~al.(2020)Westera, Mayol, and Rohde}]{westera-etal-2020-ted}
Matthijs Westera, Laia Mayol, and Hannah Rohde. 2020.
\newblock \href {https://aclanthology.org/2020.lrec-1.141} {{TED}-{Q}: {TED} talks and the questions they evoke}.
\newblock In \emph{Proceedings of the Twelfth Language Resources and Evaluation Conference}, pages 1118--1127, Marseille, France. European Language Resources Association.

\bibitem[{Wolf et~al.(2020)Wolf, Debut, Sanh, Chaumond, Delangue, Moi, Cistac, Rault, Louf, Funtowicz, Davison, Shleifer, von Platen, Ma, Jernite, Plu, Xu, Le~Scao, Gugger, Drame, Lhoest, and Rush}]{wolf-etal-2020-transformers}
Thomas Wolf, Lysandre Debut, Victor Sanh, Julien Chaumond, Clement Delangue, Anthony Moi, Pierric Cistac, Tim Rault, Remi Louf, Morgan Funtowicz, Joe Davison, Sam Shleifer, Patrick von Platen, Clara Ma, Yacine Jernite, Julien Plu, Canwen Xu, Teven Le~Scao, Sylvain Gugger, Mariama Drame, Quentin Lhoest, and Alexander Rush. 2020.
\newblock \href {https://doi.org/10.18653/v1/2020.emnlp-demos.6} {Transformers: State-of-the-art natural language processing}.
\newblock In \emph{Proceedings of the 2020 Conference on Empirical Methods in Natural Language Processing: System Demonstrations}, pages 38--45, Online. Association for Computational Linguistics.

\bibitem[{Wu et~al.(2023)Wu, Mangla, Durrett, and Li}]{wu-etal-2023-qudeval}
Yating Wu, Ritika Mangla, Greg Durrett, and Junyi~Jessy Li. 2023.
\newblock \href {https://doi.org/10.18653/v1/2023.emnlp-main.325} {{QUD}eval: {The} evaluation of questions under discussion discourse parsing}.
\newblock In \emph{Proceedings of the 2023 Conference on Empirical Methods in Natural Language Processing}, pages 5344--5363, Singapore. Association for Computational Linguistics.

\bibitem[{Yuan et~al.(2023)Yuan, Feng, Li, Wang, Pan, Wang, and Li}]{yuan2023batcheval}
Peiwen Yuan, Shaoxiong Feng, Yiwei Li, Xinglin Wang, Boyuan Pan, Heda Wang, and Kan Li. 2023.
\newblock \href {https://arxiv.org/abs/2401.00437} {Batcheval: {Towards} human-like text evaluation}.
\newblock \emph{arXiv preprint arXiv:2401.00437}.

\bibitem[{Zarcone et~al.(2016)Zarcone, Van~Schijndel, Vogels, and Demberg}]{zarcone2016salience}
Alessandra Zarcone, Marten Van~Schijndel, Jorrig Vogels, and Vera Demberg. 2016.
\newblock \href {https://www.frontiersin.org/journals/psychology/articles/10.3389/fpsyg.2016.00844/full} {Salience and attention in surprisal-based accounts of language processing}.
\newblock \emph{Frontiers in psychology}, 7:844.

\bibitem[{Zhang et~al.(2023)Zhang, Li, Ma, Zhou, and Zou}]{zhang-etal-2023-llmaaa}
Ruoyu Zhang, Yanzeng Li, Yongliang Ma, Ming Zhou, and Lei Zou. 2023.
\newblock \href {https://doi.org/10.18653/v1/2023.findings-emnlp.872} {{LLM}a{AA}: {Making} large language models as active annotators}.
\newblock In \emph{Findings of the Association for Computational Linguistics: EMNLP 2023}, pages 13088--13103, Singapore. Association for Computational Linguistics.

\bibitem[{Zhang et~al.(2020)Zhang, Kishore*, Wu*, Weinberger, and Artzi}]{Zhang2020BERTScore:}
Tianyi Zhang, Varsha Kishore*, Felix Wu*, Kilian~Q. Weinberger, and Yoav Artzi. 2020.
\newblock \href {https://openreview.net/forum?id=SkeHuCVFDr} {{BERTScore}: {Evaluating} text generation with {BERT}}.
\newblock In \emph{International Conference on Learning Representations}.

\end{thebibliography}

\clearpage
\appendix
\newpage
\section{Annotation Details} \label{appendix:details-annotation}
We use \texttt{gpt-3.5-turbo-1106} with a temperature of 0.2 for data annotation.
The prompts used for question completion (QC), question answering (QA), and question role identification (QRI) are listed in Tables \ref{tab:ann_prompt_qc}, \ref{tab:ann_prompt_qa}, and \ref{tab:ann_prompt_qri}.
\begin{table}[hbpt]
\small
\centering
\begin{tabular}{@{}p{7.5cm}@{}}
\toprule
\begin{tabular}[c]{@{}p{7.5cm}@{}}\textbf{\texttt{{[}Task Description{]}}}\\ \texttt{You will be given some questions extracted from an article and their surrounding texts. Your task is to check whether these questions are self-contained. For example, “What constitutes such a code?” is not a self-contained question due to the unclear “code”. If a question is not self-contained, complete it based on its context; otherwise, output it as it is.}\\ \\ \textbf{\texttt{{[}Input{]}}}\\ \texttt{Question 1: \{question 1\}}\\ \texttt{Context 1: \{context of question 1\}}\\ \texttt{...}\\ \\ \textbf{\texttt{{[}Output{]}}} \texttt{(Please strictly organize your output in the following format)}\\ \texttt{Question 1: \colorbox{yellow}{\textless complete question 1\textgreater }}\\ \texttt{...}\end{tabular} \\ \bottomrule
\end{tabular}
\caption{Prompt for question completion. We use the surrounding 10 sentences of a question as its context.}
\label{tab:ann_prompt_qc}
\vspace{-3mm}
\end{table}

\begin{table}[hbpt]
\centering
\small
\begin{tabular}{@{}p{7.5cm}@{}}
\toprule
\begin{tabular}[c]{@{}p{7.5cm}@{}}\textbf{\texttt{{[}Task Description{]}}}\\ \texttt{You will be given an article and some questions. In particular, these questions are posed in the article and highlighted by a marker “{[}Question{]}” before them. Your task is to answer these questions based on the article. Start by reading the article carefully and locating the given questions. For each question, check whether it is answered in the article. If not, output “no answer”; otherwise, provide an answer that should be as detailed as possible and faithful to the article. Finally, provide a confidence level from 1 to 5 for each generated answer, where 1 is the lowest and 5 is the highest.} \\ \\ \textbf{\texttt{{[}Input{]}}}\\ \texttt{Article: \{article with questions marked\}}\\ \texttt{Question 1: \{question 1\}}\\ \texttt{...}\\ \\ \textbf{\texttt{{[}Output{]}}} \texttt{(Please strictly organize your output in the following format)}\\ \texttt{Answer 1:} \colorbox{yellow}{\texttt{\textless answer to question 1\textgreater}}\\ \texttt{Confidence 1:} \colorbox{yellow}{\texttt{\textless confidence level, 1-5\textgreater}}\\ \texttt{...}\end{tabular} \\ \bottomrule
\end{tabular}
\caption{Prompt for question answering.}
\label{tab:ann_prompt_qa}
\end{table}
We ask the model to provide confidence levels for outputs of QA and QRI as these two tasks are arguably more challenging. 
437 questions have a confidence level of 1 on both tasks. 
An expert annotator (author of this work) manually validated these questions.
Note that our goal is not to annotate as many examples as possible, but to check whether there are unknown question roles beyond our taxonomy and analyze common bad cases to improve prompts.
We found that most cases with low confidence are because of their mixed roles. 
For example, when a question introduces an argument and meanwhile starts related discussions, it could be both \shading{Establish Claim} and \shading{Organize Discourse} question.
For such ambiguous cases, we recommend determining their main roles based on their question-answer relationships, as shown in Figure \ref{fig:evidence_dist}. 
For example, if the abovementioned question has an immediate answer, its main role should be \shading{Establish Claim}. 
Note that our taxonomy is to provide a holistic understanding of question functions rather than a hard classification system.

\begin{table}[t]
\centering
\small
\begin{tabular}{@{}p{7.5cm}@{}}
\toprule
\begin{tabular}[c]{@{}p{7.5cm}@{}}\textbf{\texttt{{[}Task Description{]}}}\\ \texttt{You will be given an article and some questions. In particular, these questions are presented in the article and play different roles. Your task is to identify their role. The definition and example of each question role are described below.} \\ \texttt{\{Definitions and examples of question roles\}}\\ \texttt{Make sure you understand the above instructions clearly. Start by reading the article carefully and locating the positions of the given questions. Check each question-answer pair and write down your analysis about its role based on the above definition. Finally, output its role and provide a confidence level from 1 to 5 for your judgment, where 1 is the lowest and 5 is the highest.}\\ \\ \textbf{\texttt{{[}Input{]}}}\\ \texttt{Article: \{article\}}\\ \texttt{Question 1: \{question 1\}}\\ \texttt{Answer 1: \{answer to question 1\}}\\ \texttt{…}\\ \\ \textbf{\texttt{{[}Output{]}}}\\ \texttt{Analysis 1:} \texttt{\colorbox{yellow}{\textless analysis for the role of question 1\textgreater }}\\ \texttt{Role 1:} \texttt{\colorbox{yellow}{\textless role of question 1\textgreater}}\\ \texttt{Confidence 1:} \texttt{\colorbox{yellow}{\textless confience of output, 1-5\textgreater}}\end{tabular} \\ \bottomrule
\end{tabular}
\caption{Prompt for question role identification.}
\label{tab:ann_prompt_qri}
\end{table}

\begin{table}[htbp]
\small
\centering
\begin{tabular}{@{}lcc@{}}
\toprule
\textbf{Task} & \textbf{Criteria}                 & \textbf{Yes (\%)} \\ \midrule
\textbf{QC}   & The question is self-contained    & 90.5\%             \\
\textbf{QA}   & The answer is overall acceptable  & 83.6\%              \\
\textbf{QRI}  & The identified role is correct or & 79.4\%              \\ \bottomrule
\end{tabular}
\caption{Quality review of automatic annotation.}
\label{tab:ann_quality}
\end{table}

Finally, we sample 10 research articles and 10 textbook chapters with 116 questions in total and review the accuracy of annotation. 
As shown in Table \ref{tab:ann_quality}, decent results are observed on all tasks. 
We plan to scale the dataset and update the annotation with more powerful LLMs available in the future.

\section{Experiment Details}
\subsection{Article Processing} \label{subsec:del_smo}
To process an article into the training format, we delete questions from the article and eliminate incoherence by prompting \texttt{gpt-3.5-turbo-1104} with the instruction in Table \ref{tab:filling_prompt}. 
To reduce the cost, we build the paragraph using the 5 sentences before and after the deleted question, which is enough to assess or restore the coherence of the local context according to our qualitative inspection.  
When performing this operation on random sentences, we disallow sentences around (distance<10) any already deleted or question sentences to be selected in order to avoid severe incoherence.  

\begin{table}[htbp]
\small
\centering
\begin{tabular}{@{}p{7.5cm}@{}}
\toprule

\begin{tabular}[c]{@{}p{7.5cm}@{}}\textbf{\texttt{{[}Task Description{]}}}\\ \texttt{Given a paragraph where a sentence has been removed and replaced with "{[}MASK{]}," your task is to assess whether the paragraph remains coherent without the missing sentence. If yes, simply remove the {[}MASK{]} token. If not, please edit the text around {[}MASK{]} to restore its coherence. You can only make necessary and minimal edits, leaving the majority of the paragraph verbatim. You can not introduce new information or change or remove existing information.}\\ \\ \textbf{\texttt{{[}Input{]}}}\\ \texttt{Input Paragraph: \{paragraph with a missing sentence\}}\\ \\ \textbf{\texttt{{[}Output{]}}}\\ \texttt{Coherent paragraph:}  \colorbox{yellow}{\texttt{\textless coherent paragraph\textgreater }}\end{tabular} \\ \bottomrule
\end{tabular}
\caption{Prompt for coherence maintenance.}
\label{tab:filling_prompt}
\end{table}
\subsection{Training setup} \label{subsec:training_setup}
We split the dataset into 90\% training and 10\% test sets. Our implementations are based on the Transformers Library \cite{wolf-etal-2020-transformers}.  
In concrete, for all approaches, we fine-tune the Flan-T5 for up to 10 epochs with a learning rate of $5e-5$ and batch size of 32. 
Following \citet{raffel2020exploring}, we employ the AdaFactor \cite{shazeer2018adafactor} optimizer and do not use warm-up. An early stop strategy is applied when the loss on the validation set does not decrease in three continuous epochs. We use 4 Nvidia Tesla A100 cards with 40 GB GPU memory for training. 
One epoch takes around half an hour.
At inference, we use beam search decoding with a beam size of 4. 
All evaluations are conducted with the default parameters in their public implementations.

Since the articles in our dataset are relatively long, we truncate them and keep sentences whose indices fall within $[max(0, p_0-5)$, $min(|D|, p_m+5)]$, where $|\mathcal{D}|$ is the number of sentences in article $\mathcal{D}$, and $p_0$, $p_m$ are the index of the first and last question respectively. 
In other words, we keep at least 10 sentences as the context for predicting each question. 
If the truncated document exceeds the model's context length, we split it into segments of roughly the same length and run each segment separately.

\begin{table}[tbp]
\small
\centering
\begin{tabular}{@{}p{7.5cm}@{}}
\toprule
\begin{tabular}[c]{@{}p{7.5cm}@{}}\textbf{\texttt{{[}Task Description{]}}}\\ \texttt{Given an article, your task is to incorporate several questions into the text to enhance its readability and make it more engaging.}\\ \texttt{For each question, first determine its position in the article by copying the sentence after which the question should be raised, then provide a set of keywords of the answer to the question. Finally, generate the target question.}\\ \\ \textbf{\texttt{{[}Input{]}}}\\ \texttt{\{article\}}\\ \\ \textbf{\texttt{{[}Output{]}}}\\ \texttt{Output 1:}\\ \texttt{Position:} \texttt{\colorbox{yellow}{\textless{}sentence precedes the question\textgreater}}\\ \texttt{Answer Keywords:} \colorbox{yellow}{\texttt{\textless keywords separated by ``,''\textgreater}}\\ \texttt{Question:} \texttt{\colorbox{yellow}{\textless{}question 1\textgreater}}\\ \\ \texttt{Output 2:}\\ \texttt{...}\end{tabular} \\ \bottomrule
\end{tabular}
\caption{Prompt for GPT-4 question generation}
\label{tab:gpt4_prompt}
\end{table}

\section{Human Study Details}\label{A:details-human-study}
\subsection{Demographic Information of Participants}\label{subsec:demo_info}
The average age was 29 years, with 25 female and 20 male participants. 
The simplified ethnicity distribution is: 23 white, 15 black, and 7 Asian. 
All information is on a self-identification basis.

\begin{figure*}[htbp]
\centering
\includegraphics[width=0.9\linewidth]{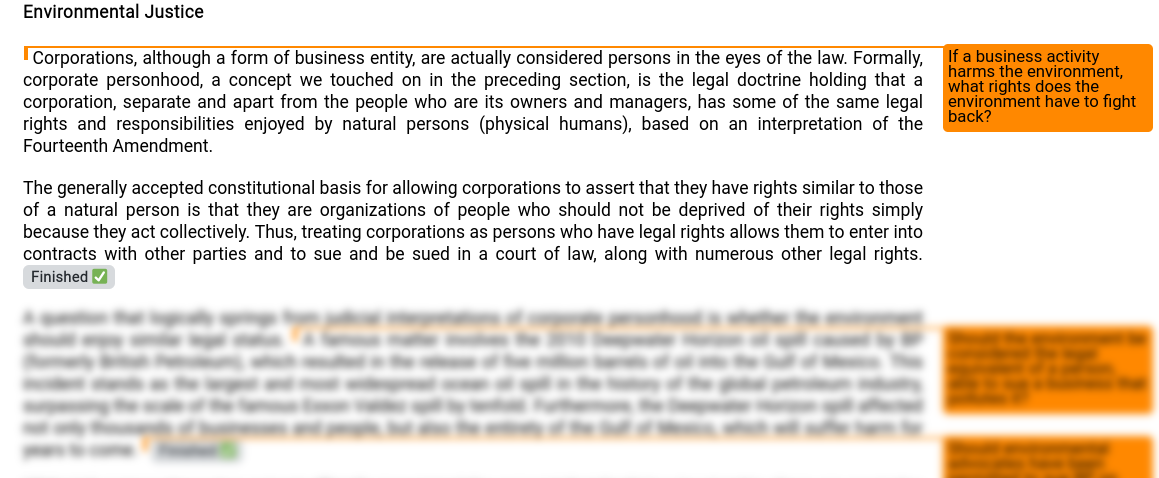}
\caption{Main reading interface. We highlight reference or generated questions on the right side. In order to measure paragraph-level reading time, participants have to click \textit{Finished} at the end of each paragraph in order to reveal the next one.}
\label{fig:main-reading-screenshot}
\end{figure*}
\subsection{Reading Interface}\label{sub:app_read_interface}
Figure \ref{fig:main-reading-screenshot} shows a screenshot of the reading interface. 
To highlight guiding questions, we display them on the right side of the article. 
In order to measure paragraph-level reading time, participants need to click the "\texttt{Finished}" button at the end of each paragraph to reveal the next one.

\subsection{Summary Evaluation}\label{sec:app_sum_eval}
We prompt GPT-4 (\texttt{gpt4-1104-preview}) with a temperature of 0.7 to evaluate the quality of collected summaries.
The prompt is shown in Table \ref{sec:app_sum_eval}, where we adopt the chain-of-thought \cite{NEURIPS2022_9d560961} and batch evaluation \cite{yuan2023batcheval} prompting strategy.  

\subsection{Additional Summary Analysis}\label{subsec:human_add_result}

\begin{table}[t]
\centering
\resizebox{\linewidth}{!}{
\begin{tabular}{lp{6.7cm}}
\toprule
Engaging & The questions helped in keeping me \textbf{engaged} with the text. \\
Understanding & The questions improved my \textbf{understanding} of the structure and main ideas of the article. \\
Overall & \textbf{Overall}, I prefer to have such questions during the reading. \\
\cmidrule{1-2}
Relevant & Is the question is \textbf{relevant} to the context? \\
Position & Is the question raised at an \textbf{appropriate position} and not distracting? \\
Important & Is the question \textbf{important} to the central topic of the article? \\
\bottomrule
\end{tabular}
}
\caption{Evaluation questions used in the human study.}
\label{tab:human_eval_questions}
\vspace{-2mm}
\end{table}
We provide additional analysis about collected summaries in Table \ref{tab:add_summary}, including summary time, length, and n-gram overlap between summaries and their sources.
We can observe the reference and generated group produced longer summaries than the control group and spent longer time accordingly. 
The authentic group's summaries and the generated group's summaries are similar in quality, but the former is more concise.
This can be attributed to the greater significance of reference questions, consistent with user ratings in Table \ref{tab:human-quantitative}. 
In addition, summaries from both the reference and specifically generated groups show a larger n-gram overlap, providing further evidence of improved memorization.

\begin{table}[t]
\centering
\small
\setlength\tabcolsep{4pt}
\begin{tabular}{@{}lccccc@{}}
\toprule
\multirow{2}{*}{} & \multicolumn{1}{l}{\multirow{2}{*}{\textbf{\begin{tabular}[c]{@{}l@{}}Sum.\\ Length\end{tabular}}}} & \multicolumn{1}{l}{\multirow{2}{*}{\textbf{\begin{tabular}[c]{@{}l@{}}Sum. \\ Time\end{tabular}}}} & \multicolumn{3}{c}{\textbf{N-gram Overlap (\%)}}                                                                          \\ \cmidrule(l){4-6} 
                  & \multicolumn{1}{l}{}                                                                                & \multicolumn{1}{l}{}                                                                               & \multicolumn{1}{l}{\textbf{uni-gram}} & \multicolumn{1}{l}{\textbf{bi-gram}} & \multicolumn{1}{l}{\textbf{tri-gram}} \\ \midrule
\textbf{Contr.}   & 117                                                                                                 & 619                                                                                                & 70.8                                  & 6.22                                 & 0.69                                  \\
\textbf{Auth.}    & 123                                                                                                 & 627                                                                                                & 74.1                                  & 6.45                                 & 0.71                                  \\
\textbf{Gen.}     & 146                                                                                                 & 745                                                                                                &\textbf{76.0}                                  & \textbf{8.84}                                 & \textbf{1.44}                                  \\ \bottomrule
\end{tabular}
\caption{Averaged summary length (words), summarization time (seconds), and n-gram overlap between summaries and articles.}
\label{tab:add_summary}
\vspace{-5mm}
\end{table}

\begin{table*}[htbp]
\small
\centering
\begin{tabular}{@{}p{15cm}@{}}
\toprule
\multicolumn{1}{c}{\textbf{Prompt Template}}                                                                                                                                                                                                                                                                                                                                                                                                                                                                                                                                                                                                                                                                                                                                                                                                                                                                                                                                                                                                                                                                                                                                                                                                                                                                                                                                                                                                                                     \\ \midrule
\begin{tabular}[c]{@{}p{15cm}@{}}\textbf{\texttt{{[}Task Description{]}}}\\ \texttt{Your will be given three summaries written for an article titled “\{title\}”. Please act as an impartial judge and evaluate the quality of these summaries in terms of \{metric\}. Please make sure you read and understand the following instructions carefully. Please keep this document open while reviewing and refer to it as needed.}\\ \\ \textbf{\texttt{{[}Evaluation Criteria{]}}}\\ \texttt{\{metric description\}}\\ \\ \textbf{\texttt{{[}Evaluation Steps{]}}}\\ \texttt{Read the source article carefully and identify the main topic and key points.}\\ \texttt{Read each summary carefully. Check if the summary meets the above criteria and provide an explanation for your judgment.}\\ \texttt{Assign a \{metric\} score for each summary on a scale of 1 (lowest) to 5 (highest). Decimal scores are particularly encouraged.}\\ \\ \textbf{\texttt{{[}Input{]}}}\\ \texttt{Article: \{article\}}\\ \texttt{Summary 1: \{summary\_1\}}\\ \texttt{Summary 2: \{summary\_2\}}\\ \texttt{Summary 3: \{summary\_3\}}\\ \\ \textbf{\texttt{{[}Output{]}}} \texttt{(Please strictly organize your output in the following format)}\\ \texttt{Explanations:}\\ \texttt{Analysis of summary 1:} \colorbox{yellow}{\texttt{\textless{}your analysis\textgreater}}\\ \texttt{Analysis of summary 2:} \colorbox{yellow}{\texttt{\textless{}your analysis\textgreater}} \\ \texttt{Analysis of summary 3:} \colorbox{yellow}{\texttt{\textless{}your analysis\textgreater}} \\ \\ \texttt{Scores:}\\ \texttt{Score for summary 1:} \colorbox{yellow}{\texttt{\textless{}score only, 1 - 5, preferably decimal\textgreater}}\\ \texttt{Score for summary 2:} \colorbox{yellow}{\texttt{\textless{}score only, 1 - 5, preferably decimal\textgreater}}\\ \texttt{Score for summary 3:} \colorbox{yellow}{\texttt{\textless{}score only, 1 - 5, preferably decimal\textgreater{}}}\end{tabular} \\ \midrule
\multicolumn{1}{c}{\textbf{\{metric description\}}}                                                                                                                                                                                                                                                                                                                                                                                                                                                                                                                                                                                                                                                                                                                                                                                                                                                                                                                                                                                                                                                                                                                                                                                                                                                                                                                                                                                                                              \\ \midrule
\begin{tabular}[c]{@{}p{15cm}@{}}\texttt{\textbf{Coherence} (1-5) - the collective quality of all sentences. The summary is well-structured and well-organized. The summary should not just be a heap of related information, but should build from sentence to sentence to a coherent body of information about a topic.}\\ \texttt{\textbf{Consistency} (1-5) - the factual and conceptual alignment between the summary and the source article. The summary faithfully and precisely conveys the messages and ideas from the source material, without distortion or misinterpretation.}\\ \texttt{\textbf{Informativeness} (1-5): the extent to which the summary encapsulates the essential and relevant information. Informativeness is not merely about including various pieces of information, but selecting the most crucial elements that offer a comprehensive understanding of the topic.}\end{tabular}                                                                                                                                                                                                                                                                                                                                                                                                                                                                                                                                                                                                                          \\ \bottomrule
\end{tabular}
\caption{Prompt template and metric descriptions for summary evaluation.}
\label{tab:sum_eval_prompt}
\end{table*}

\subsection{Observations from Preliminary Study} \label{Appendix: observation}
Before the reported human study, we conducted preliminary interviews and follow-up surveys with 15 participants, five from each group, as detailed in Sec.~\ref{subsec: exp_design}.
The aim is to validate the experiment design and gather user preferences and feedback on presenting questions during their article reading experience. 
We refer to participants as \texttt{group\_id} (R/C/G) + \texttt{user\_id} (1-5), where R, G, and C stand for Reference, Generated, and Controlled. 

Many participants expressed positive feedback on the generated questions, noting that ``\textit{the questions are easy to understand}'' (G5). 
Notably, compared to the author-curated questions with relatively complex terms and sentences, our generated questions had simpler vocabulary and shorter sentences and facilitated quicker context comprehension.
Some participants even found it helpful to use the questions to ``\textit{remember the content of the whole article}'' and ``\textit{better understand it again}'' (G1).
We summarize the observations from the interview into three points.
\paragraph{Consistent Question Preference.}
While each participant has their unique criteria for defining a good question, we have observed a consistent preference for questions that are both intriguing and informative. 
Among the 10 participants exposed to supported questions, whether authentic or generated, seven conveyed a preference for questions that stimulate thought and reflection.
These questions may offer insights (R5) or highlight specific details (G1), but they must be inherently challenging and motivate reading (G5).
Conversely, participants tend to dislike questions that are too easy or have immediate answers within the article (R4, R5, G5). 
This observation provides valuable insights for refining our future question generation process by allowing us to better control the level of difficulty and align with participants' preferences.

\vspace{2mm}

\noindent \textbf{Relevance, Distractibility, and Helpfulness.}
In the survey, we asked participants to assess the relevance, distractibility, and helpfulness of each question on a 5-point Likert scale.
Half believe a question's relevance depends on whether the following sentences answer it.
Meanwhile, shorter questions tend to be less distracting, with their distractibility rating inversely proportional to perceived helpfulness. 
Interestingly, conflicting viewpoints arose, with some participants considering certain questions neither distracting but useless (R1) or helpful yet irrelevant (G1, G4). 
This nuanced understanding of human complexity calls for more research in human-AI collaboration research to find adaptive question generation solutions.

\vspace{2mm}
\noindent \textbf{Question Position and Scenario Matters.}
The two observations mentioned above may vary slightly depending on the question's position and the reading scenario. 
Notably, five participants experienced a ``cold start'' during the reading task, expressing difficulty in ``\textit{getting into the context of an article at first}'' (R1). 
Therefore, their initial preference leans towards easier and more relevant questions at the article's beginning. 
As they familiarize themselves with the context after reading a few paragraphs, their inclination shifts towards more intriguing and divergent questions.
Furthermore, participants' standards for a good question may fluctuate in different scenarios. 
Distinctions were made between first-time reading versus content reviewing (R2), learning scenarios versus examination scenarios (R4), and serious learning versus casual reading (G4). 
These considerations could serve as additional factors in our future iterations.

\end{document}